\documentclass[11pt,a4paper]{article}
\usepackage{times,latexsym}
\usepackage{url}
\usepackage[T1]{fontenc}

\usepackage[acceptedWithA]{tacl2021v1}
\usepackage{xspace,mfirstuc,tabulary}

\newif\iftaclinstructions
\taclinstructionsfalse % AUTHORS: do NOT set this to true
\iftaclinstructions
\renewcommand{\confidential}{}
\renewcommand{\anonsubtext}{(No author info supplied here, for consistency with
TACL-submission anonymization requirements)}
\newcommand{\instr}
\fi

\iftaclpubformat % this "if" is set by the choice of options

\else

\fi

\newcommand{\Mname}{CLD}
\renewcommand{\cite}{\citep}
\usepackage{amssymb}
\usepackage{amsmath}
\usepackage{todonotes}
\usepackage{booktabs}
\usepackage{multirow}
\usepackage{wrapfig}

\usepackage{makecell}
\begin{document}

%\title{Contrastive-Learning-based Disentanglement via\\ Imitating the Attribute Control Process}
\title{Text Attribute Control via Closed-Loop Disentanglement}
\author{
  Lei Sha$^{1,2,4}$ 
  \and
  Thomas Lukasiewicz$^{3,2}$
  \\
  $^1$Institute of Artificial Intelligence, Beihang University\\
  $^2$Department of Computer Science, University of Oxford, UK \\
  $^3$Institute of Logic and Computation, Vienna University of Technology, Austria\\
  $^4$ Zhongguancun Laboratory, Beijing, China\\
\texttt{shalei@buaa.edu.cn}, \quad\texttt{thomas.lukasiewicz@tuwien.ac.at}
}

\maketitle
\begin{abstract}
Changing an attribute of a text without changing the content  usually requires to first disentangle the text into irrelevant attributes and content representations. After that,  in the inference phase,   the representation of one attribute is tuned to a different value, expecting that the corresponding attribute of the text can also be changed accordingly. The usual way of disentanglement is to add some constraints on the latent space of an encoder-decoder architecture, including adversarial-based constraints and mutual-information-based constraints.
However, the previous semi-supervised processes of attribute change are usually not enough to guarantee the success of attribute change and content preservation.
In this paper, we propose a novel approach to achieve a robust control of attributes while enhancing content preservation. 
In this approach, we use a semi-supervised contrastive learning method to encourage the disentanglement of attributes in latent spaces. Differently from previous works, we re-disentangle the reconstructed sentence and compare the re-disentangled latent space with the original latent space, which makes a closed-loop disentanglement process.  This also helps content preservation. 
In addition,  the contrastive learning method is also able to replace the role of minimizing mutual information and adversarial training in the disentanglement process, which alleviates the computation cost. We conducted experiments on three text datasets, including the Yelp Service review dataset, the Amazon Product review dataset, and the GoEmotions dataset. The experimental results show the effectiveness of our model.

\end{abstract}

\section{Introduction}
Controlling the attributes of a text is an important application of interpretable natural language models. The term  ``control'' usually means to take attributes as a handle, and pulling the handle causes corresponding changes in the text. The control process should not change the content of the text. Usually, this is realized by disentangling the text into  multiple irrelevant latent spaces for content and multiple attributes~\cite{sha2021multi}.

Previous works mainly use two methods for disentangling the attributes: adversarial learning~\cite{chen2016infogan,john-etal-2019-disentangled} and mutual information minimization~\cite{moyer2018invariant,sha2021multi}. For each latent space (corresponding to the content or attributes), the former~\cite{john-etal-2019-disentangled} applies adversarial training  to reduce  the  information that should not be contained in that space. Also, \newcite{logeswaran2018content} uses an adversarial method to encourage the generated text to be compatible with the tuned attributes. To alleviate the training cost and the instability of adversarial methods, \newcite{moyer2018invariant} and \newcite{sha2021multi} proposed to minimize the mutual information between different latent spaces.

%However, previous disentanglement approaches did not pay too much attention on  ensuring  the success of attribute transferring. To be more specific, 
\begin{figure}[!t]
    \centering
    \includegraphics[width=\linewidth]{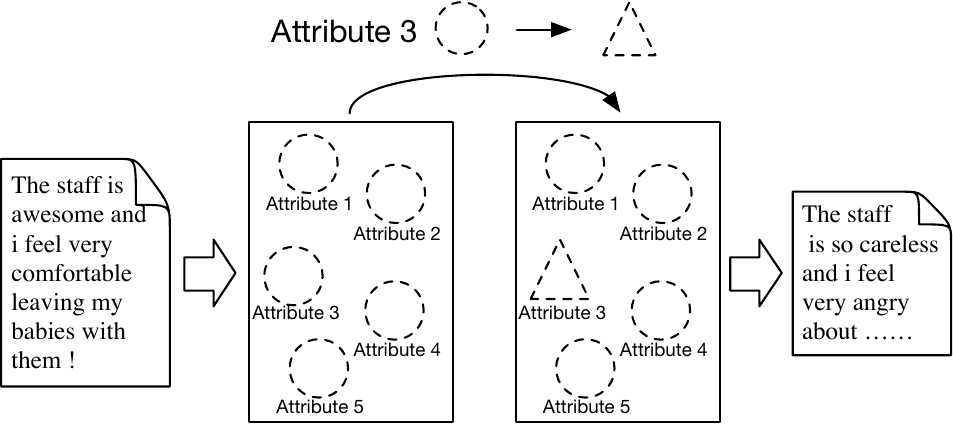}\\
    \caption{Attribute control: a sentence is disentangled into separate attributes. Each dashed circle represents an attribute. After one of the attributes was changed to another value (here, attribute 3 was changed from a circle to a triangle), the corresponding attribute of the reconstructed sentence was changed accordingly.}
    \label{fig:intro}
\end{figure}

When changing attributes, previous methods change the representation of an attribute in the latent space, expecting the generated text to satisfy the changed attribute. However,  the generated text  does not necessarily do so and preserve the content as well as other attributes, if this is not explicitly encouraged  in the training process.

\begin{figure*}
    \centering
    \includegraphics[width=0.8\linewidth]{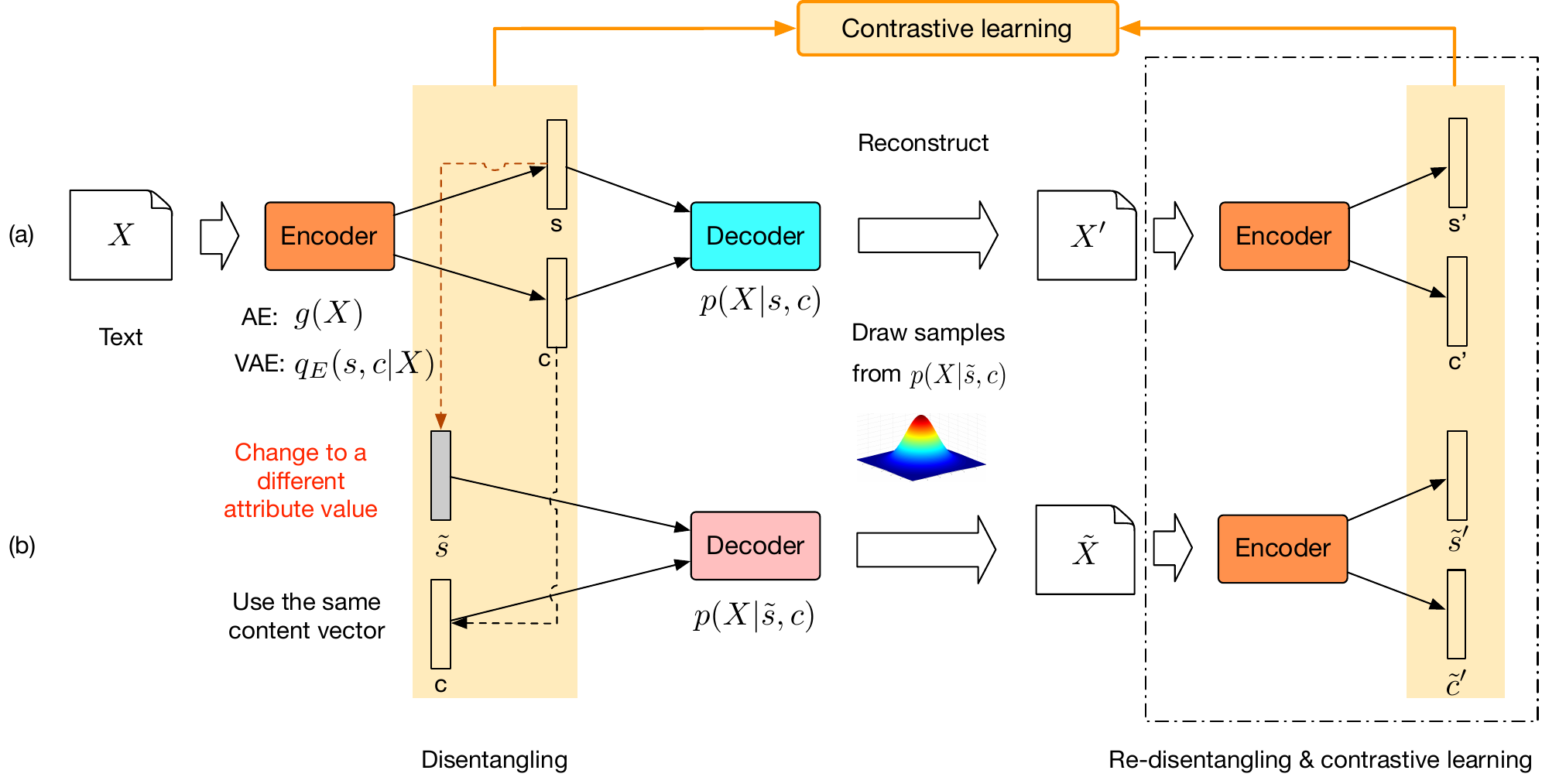}
    \caption{Complete architecture of our proposed model \Mname. The upper row (a) represents the normal disentanglement process. The lower row (b) imitates the style/attribute transfer process. In both  processes, we conduct re-disentanglement and use contrastive learning  to encourage the content vector ($c$) to stay unchanged, while the style vectors ($s$, $\tilde s$) change to the desired values. }
    \label{fig:arch}
\end{figure*}

In this paper, we propose   a novel attribute control model, which uses contrastive learning to make the   latent representation of attributes irrelevant to each other, while encouraging the content to be unchanged during attribute control.
We still use an autoencoder architecture to disentangle the text into  latent spaces. Inspired by closed-loop control systems~\cite{di1967feedback} and closed-loop data transcription~\cite{dai2021closed}, we utilize the encoder once more to disentangle the generated text into re-disentangled latent spaces. This enables the disentanglement process to operate in a closed-loop manner, resulting in greater stability. Then, we use contrastive learning to reduce the difference of unchanged attributes between the original and the re-disentangled latent spaces, while enlarging the difference between changed attributes. %between cases? inside one case?
The contrastive learning method thus  provides an alternative way for disentanglement, since it directly encourages content preservation and non-target attribute preservation when changing the targeted attribute.

Our contributions are briefly summarized as follows:
\begin{itemize}%[topsep=0pt,parsep=0pt,partopsep=0pt]
    \item We propose a new approach to disentanglement based on contrastive learning, where we re-disentangle the reconstructed sentence and compare the re-disentangled latent space with the original latent space to make a closed-loop control.

    % use contrastive learning to encourage  the generated text to be   compatible to the transferred attribute.
    \item We propose several contrastive learning loss functions  to   disentangle the text into irrelevant latent spaces as a   replacement for adversarial learning or mutual information minimization.
    \item We conduct extensive experiments on three text datasets (Yelp Service review, Amazon Product review, and GoEmotions dataset) to show the disentanglement effectiveness of our method.
\end{itemize}

\section{Related Works}
\paragraph{Disentanglement for Attribute Control.}
For a natural text, if we want to change one of its attributes while keeping all its other attributes unchanged, a promising way is to disentangle the attributes from the text. Then, changing one attribute is  not expected to affect other attributes. % so that the controlling of the attributes via the disentangled representations is conducted. %TODO more profitable to occur in the intro?

Techniques for disentangling attributes can be divided into two different types: explicit disentanglement~\cite{chen2016infogan,john-etal-2019-disentangled,sha2021multi} and implicit disentanglement~\cite{higgins2017beta,chen2018isolating}. Explicit disentanglement requires  the training dataset to contain attribute annotations, which may help to separate the latent space into interpretable components for each attribute. For example, \newcite{chen2016infogan} and \newcite{john-etal-2019-disentangled} used adversarial methods to reduce the influence between latent spaces. To overcome the training difficulties and resource-consuming problems of adversarial methods, mutual information minimization methods~\cite{moyer2018invariant,sha2021multi} have been proposed  to conduct disentanglement in a non-adversarial way.
The explicit disentanglement method is easier for attribute control,  because it is easy to tell the model which part of the latent space represents which  attribute. 

Implicit disentanglement does not use the attribute annotations in the training dataset, so for each disentangled component, it is hard to tell exactly which attribute it corresponds to. Implicit disentanglement includes $\beta$-VAE~\cite{higgins2017beta}, $\beta$-TCVAE~\cite{chen2018isolating}, and many derivatives~\cite{mathieu2018disentangling,kumar2017variational,esmaeili2018structured,hoffman2016elbo,narayanaswamy2017learning,kim2018disentangling,shao2020controlvae}. The basic principle of implicit disentanglement is to capture the internal relationship between input examples. For example, \newcite{chen2018isolating} break the  evidence lower bound~(ELBO) into several parts and proposed the \emph{Total Correlation}, which  encourages the different  attributes to be statistically independent.  \emph{Total Correlation} is also the cornerstone for MTDNA~\cite{sha2021multi}. \newcite{esmaeili2018structured} further break the ELBO into more segments and discussed the effect of each segment toward implicit disentanglement. However, without the help of annotation, it is difficult for implicit disentanglement to obtain better disentangled latent spaces than  explicit disentanglement.

\paragraph{Attribute Control without Disentanglement.}
Although disentanglement is a general way to perform attribute control, there are also methods that control attributes without disentanglement. For example, \newcite{logeswaran2018content} use adversarial training to  judge  whether the generated sentence is compatible with the target attribute label. \newcite{lample2019multipleattribute}  use a back translation method to model the attribute control process. Similar methods are also applied by \newcite{Luo2019TowardsFT}, \newcite{artetxe2018unsupervised}, and \newcite{artetxe2019effective}. Other methods also tried some other task formulations, like probabilistic inference by HMM~\cite{he2019probabilistic} and paraphrase generation~\cite{krishna2020reformulating}.

\paragraph{Contrastive Learning.}
Contrastive learning has been proposed by \newcite{hadsell2006dimensionality}, and has witnessed a series of developments in recent years. The goal of contrastive learning can be seen as training an encoder for a dictionary look-up task~\cite{he2020momentum}. 
% The loss function of contrastive learning  
Triplet loss~\cite{chechik2010large,hoffer2015deep,wang2015unsupervised,sermanet2018time} has originally been proposed to achieve this, which reduces the distance between the example and a positive example and enlarges the distance between the example and a negative example. Noise contrastive estimation (NCE) loss~\cite{gutmann2010noise,gutmann2012noise} uses a probabilistic model to discriminate the positive and negative examples. Based on NCE,  InfoNCE~\cite{oord2018representation,hjelm2018learning,anand2019unsupervised,bachman2019learning,gordon2020watching,hjelm2020representation,zhuang2019local,xie2020delving,khosla2020supervised}  has a similar form of classification-based N-pair loss~\cite{le2020contrastive}, and it has proved that the minimization of InfoNCE also maximises the lower bound of the mutual information between the input and the representation~\cite{oord2018representation}. Similar mutual-information-based losses include DIM~\cite{hjelm2018learning}, PCL~\cite{li2020prototypical}, and SwAV~\cite{caron2020unsupervised}. Also, MoCo~\cite{he2020momentum,chen2020improved,chen2021empirical} uses a dynamic memory queue for building large and consistent dictionaries  for unsupervised learning with InfoNCE loss. SimCLR~\cite{chen2020simple,chen2020big} uses a large batch size in an  instance discrimination task.

In contrast to the above, instead of on the input examples, we apply contrastive learning in the original and re-disentangled latent spaces to encourage that attributes can be robustly controlled, which thus makes the latent space disentangled. To our knowledge, this is the first work of using contrastive learning in such a way to conduct disentanglement. %Although InfoNCE~\cite{oord2018representation} is s

\paragraph{The difference between our approach and other disentanglement methods.}
Our \Mname\   exploits the essence of attribute disentanglement. We now compare it with two previous methods of  disentanglement.

Adversarial disentanglement~\cite{chen2016infogan,john-etal-2019-disentangled} naturally uses adversarial methods to eliminate the information of other attributes from the representation of one attribute. However, if there are multiple style types, then we need one discriminator for each of the style types, which is a massive cost of resources. Also, adversarial methods can only be taken as constraints on the latent space, since they do not directly encourage the other attributes not being affected by the changed attribute.

Another method is mutual information minimization~\cite{moyer2018invariant,sha2021multi}, which is more efficient and elegant. However, it still does not directly encourage that the change in the style's latent space can be perfectly reflected in the output sentence.  On the other hand, it is based on some strong assumptions like that the content vector should also follow a Gaussian distribution. But in our  \Mname, the contrastive-learning-based method does not require any of these assumptions. Moreover, \Mname\ directly models the attribute control process in an easier and more natural way, which is more flexible to be generalized to more complex attributes and  latent spaces.

\vspace*{-1ex}\section{Approach}
In this section, we introduce the design of our model
for \emph{contrastive learning disentanglement (CLD)}. Differently from  previous works, our proposed model is very simple, as it only contains the basic encoder-decoder architecture and three contrastive learning loss functions. The architecture of our model is shown in Figure~\ref{fig:arch}.

\subsection{Basic Architecture for Disentanglement}
Like previous disentanglement methods~\cite{higgins2017beta,john-etal-2019-disentangled,sha2021multi}, we use an autoencoder as our basic architecture. Autoencoders are able to map the input text into a latent space, while encouraging the latent vector to contain the complete information of the input. So,  disentanglement is usually achieved by adding constraints to  the   latent space to  split it into irrelevant segments. Then, each segment represents an isolated feature of the input, and once changed the reconstructed text should also be changed correspondingly.

For explicit disentanglement (with annotated attributes for training), we use two kinds of autoencoders: vanilla autoencoders~\cite{hinton1994autoencoders} and variational autoencoders (VAEs)~\cite{kingma2014auto}. Given a text dataset $S_X=\{X_1, \ldots, X_N\}$, the loss functions of these two autoencoders are defined as follows:
\begin{align}
  J_\text{AE} =& -\mathbb E_{X\sim S_X} p(X|f(X)),\\
  J_\text{VAE} =&-\mathbb E_{z\sim q_E(z|X)}\log [p(X|z)] \nonumber\\
  +& \lambda_\text{KL}\text{KL}(q_E(z|X)||p(z))\label{eq:vae},
\end{align}
where $f(\cdot)$ and $q_E(z|X)$  are the encoders in the vanilla  and the variational autoencoders, respectively, $p(X|z)$ is the  decoder, and $p(z)$ is a prior distribution (usually, $\mathcal N(0,1)$).
The detailed architecture %of the encoder and the decoder  
is given in the appendix.

Note that $J_{\text{VAE}}$ has the name "VAE" because the latent space is calculated using the same method as a variational autoencoder (VAE). Specifically, a VAE uses an encoder to generate a distribution over the latent space, and then samples a vector $z$ from this distribution, and then feeds $z$ to a decoder. Sampling from a distribution results in a continuous latent space~\cite{bowman2016generating}.

%For implicit disentanglement (with no annotated attributes for training), we only use  variational auto-encoders since they are proved to be capable of disentangling latent spaces without priori knowledge of the data~\cite{higgins2017beta,chen2018isolating}.

\subsection{Contrastive Learning for Explicit Disentanglement}

Contrastive learning is originally proposed to learn such an embedding space in which similar sample pairs stay close to each other, while dissimilar ones are far apart. So, for disentangled representations, we can re-disentangle the reconstructed input and conduct contrastive learning between the disentangled representations and re-disentangled representations. Intuitively, after one disentangled feature is changed, the corresponding re-disentangled feature should also be changed, and the other re-disentangled features should remain unchanged.

\paragraph{Basics for Explicit Disentanglement.}
In explicit disentanglement, the most typical way is to separate the latent space into two irrelevant latent spaces, one for the style ($s$) and one for the content ($c$)~\cite{john-etal-2019-disentangled,sha2021multi}. 
The style\footnote{Following the glossary by \newcite{sha2021multi}, a \textbf{style type} is a style class that represents a
specific feature of text or an image, e.g., sentiment, tense, or
face direction; and a \textbf{style value} is one of the different values within a style type, e.g., sentiment (positive/negative), or
tense (past/now/future). }  vector here is the representation of one of the attributes of the text, including  sentiment, tense, and tone for text. In this paper, we define a new symbol to represent the disentanglement: ``$\twoheadrightarrow$''. Then, $X\twoheadrightarrow [s, c]$ represents that the representations of $s$ and $c$ are obtained by directly splitting the latent vector $z$ (in Eq.~\eqref{eq:vae}) into $s$ and $c$. On the other hand, we  define ``$\rightarrowtail$'' for generating text according to the disentangled attributes. 
By Eq.~\eqref{eq:vae}, the distribution of the generated text is calculated by $p(X|s,c)$. So, we can take a sample text from this distribution as  $X' \sim p(X|s,c)$, which is denoted by $[s, c]\rightarrowtail X'$ in this paper.
Then, the disentanglement process and the reconstruction process are written  as:
\begin{equation}\label{eq:recon_ori}
    X \twoheadrightarrow [s, c], \qquad [s, c]\rightarrowtail X',
\end{equation}
where $X'$ represents the reconstructed text.

\paragraph{Re-disentanglement for Style Transfer.}
Following the unified distribution-control (UDC) method  in \cite{sha2021multi}, we also predefine a Gaussian distribution $\mathcal N_i$ for the $i$-th style type value. To give a specific example,  there are two values for text sentiment (positive and negative), each corresponds to a Gaussian distribution. 

To directly model the style transfer process,  we first change the style vector $s$ to the vector of a different style, which is sampled from the  unified style distribution defined by the UDC method. In the training phase, this sampling process can be conducted by the reparameterization trick as shown in \cite{kingma2014auto}.
Then, we   reconstruct the text 
and disentangle the text for a second time (namely, re-disentangle) into style vector and content vector.

In detail, assuming that there are $V$ possible style values for  $s$, we sample $v$  style values  $\tilde{s}_1,\ldots,\tilde{s}_v$ that are different from $s$'s original style value. Then, we  replace $s$ with $\tilde{s}$\footnote{The subscript is omitted, since we do the same operation for each style type value sample.} and generate the text $\tilde{X}$.
After that, we re-disentangle the  generated text $X'$ (in Eq.~\eqref{eq:recon_ori}) and $\tilde{X}$, and compare the re-disentangled representation of style  and content   with the  original representation of style  and content.

So, the generation and re-disentanglement process can be described as follows:
% \begin{equation}
\begin{align}
    &[s, c] \rightarrowtail X', \quad  X' \twoheadrightarrow [s', c'];\\
    &[\tilde s,c]\rightarrowtail\tilde X,  \quad\tilde X \twoheadrightarrow [\tilde s', \tilde c']. \label{eq:transfer_dis}
\end{align}
% \end{equation}

\paragraph{Contrastive Learning.}\label{sec:edcl}
First, under the UDC setting, assume that the predefined trainable distributions for each style value are $\mathcal N_1, \ldots, \mathcal N_V$. The disentangled style vector $s$ is expected to be close to the corresponding style value's representation $s^\text{pre}_*$\footnote{$s^\text{pre}_i$ is sampled from the distribution $\mathcal N_i$. $s^\text{pre}_*$ is sampled from the distribution $\mathcal N_*$, which corresponds to  the ground truth attribute label. } and far away from other style values' representation.  Consistent with previous works~\cite{he2020momentum}, we use the dot product to measure the similarity and the InfoNCE~\cite{oord2018representation} loss as the contrastive learning loss function as follows:
\begin{align}
    &s^\text{pre}_i \sim \mathcal N_i, \quad i\in \{1,\ldots, v\}, \\
    & L_\text{ori} = -\log\frac{\exp(s\cdot s^\text{pre}_*/\tau)}{\sum_{i=0}^v \exp(s\cdot s^\text{pre}_i/\tau)},
\end{align}
where $\tau$ is a temperature hyperparameter \cite{he2020momentum}.

When we re-disentangle the reconstructed text as $X'\rightarrowtail [s',c']$, the representation for style $s'$ should be close to the original style value, and far away from all the other style value's representations. %The  content's representation $c'$ should  be close to $c$.  
The corresponding InfoNCE~\cite{oord2018representation} loss is as follows:
\begin{align}
    L_\text{re} = -\log\frac{\exp(s'\cdot s_*^\text{pre}/\tau)}{\sum_{i=0}^v \exp(s'\cdot s_i^\text{pre}/\tau)}.
\end{align}
On the other hand, when the style transfer process is conducted as Eq.~\eqref{eq:transfer_dis}, ideally,   the re-disentangled style representation   $\tilde s'$ should be far from the original style $s$ and close to the transferred style vector $\tilde s$. %Also, the  content's representation $\tilde c'$ should  be close to $c$.
So, the InfoNCE~\cite{oord2018representation} loss function for each of the sampled style values, namely, $\tilde s_k$ ($k=1,\ldots,v$), is as follows:
\begin{equation}\label{eq:7}
\begin{small}
{\tilde L_k=-\log\frac{\exp(\tilde s'_k\cdot \tilde s_k/\tau)}{\exp(\tilde s'_k\cdot s/\tau)+\sum_{i=0,i\ne k}^v \exp(\tilde s'_k\cdot \tilde s_i/\tau)},}
\end{small}
\end{equation}
For the re-disentangled content representations $c'$ and $\tilde c'$, it should be close to the original content representation $c$ and far from the content representation disentangled from other examples. The InfoNCE loss for content representation is $L_c(c')$. Similarly, the contrastive learning constraint for $\tilde c'$ is  $L_c(\tilde c')$ as follows.
\begin{align}
L_c(c') &= -\log\frac{\exp(c'\cdot c/\tau)}{\sum_{i=0}^M \exp(c'\cdot c^{(i)}/\tau)},\\
 L_c(\tilde c') &= -\log\frac{\exp(\tilde c'\cdot c/\tau)}{\sum_{i=0}^M \exp(\tilde c'\cdot c^{(i)}/\tau)},
\end{align}
where $c^{(i)}$ is the disentangled content representation of the $i$-th example in the current batch, $M$ represents the batch size.

% Besides, a similar contrastive learning loss  should also be applied on the original content vector as follows:
% \begin{align}\label{eq:10}
% L_c(c) = -\log\frac{\exp(c\cdot c/\tau)}{\sum_{i=0}^M \exp(c\cdot c^{(i)}/\tau)}.
% \end{align}
% So, the input examples with different semantic meanings are much more distinguishable in the content space, which makes it easier for encouraging content preservation by contrastive learning. 
\begin{table*}
\resizebox{\linewidth}{!}{
    %\begin{tabular}{l|p{5cm}|p{1.5cm}|p{2.7cm}|c|p{1.5cm}|p{2.5cm}}
    \begin{tabular}{l|c|c|c|c|c|c}
    \toprule[1.0pt]
      Sentiment:& positive  & \multicolumn{4}{c|}{negative} & ambiguous \\
    \midrule[0.5pt]
      Ekman: & joy & fear&sadness&disgust&anger & surprise\\
    \midrule[0.5pt]
      Emotions:& \makecell[l]{joy, amusement, approval, \\excitement, gratitude,  love,\\ optimism, relief, pride, \\admiration, desire, caring}  & \makecell[l]{fear, \\nervousness} &\makecell[l]{disappointment, \\embarrassment,\\sadness, grief,\\ remorse}& disgust & \makecell[l]{anger, \\annoyance, \\disapproval}& \makecell[l]{surprise, \\realization, \\confusion, \\curiosity}\\
    \bottomrule[1.0pt]
    \end{tabular}
    }
   \caption{Mapping of Emotion Categories to Sentiment and Ekman Taxonomy in GoEmotions Dataset}
   \label{tab:goemo}
\end{table*}

Finally, if we are using a vanilla autoencoder as the basic architecture, the total loss function of contrastive-learning-based explicit disentanglement is shown in Eq.~\eqref{eq:ttl_ed}.
\begin{equation}\label{eq:ttl_ed}
    L=J_\text{AE}+\lambda_\text{ori}L_\text{ori} +\lambda_\text{re}L_\text{re} + \lambda_{k}\sum\nolimits_{k=1}^v\tilde L_k + \lambda_cL_c,
\end{equation}
where $\lambda_\text{ori}$, $\lambda_\text{re}$, $\lambda_{k}$, and $\lambda_c$ are hyperparameters. When we are using a VAE as the basic architecture, we only need to replace $J_\text{AE}$ with $J_\text{VAE}$ in Eq.~\eqref{eq:ttl_ed}. $L_c$ is obtained by summing up the two contrastive learning losses for content preservation as shown in Eq.~\eqref{eq:lc}. The coefficients of the three items are set to the same, because they are expected to provide an equal effect on the three latent spaces: the original latent space, the re-disentangled latent space, and the style-transferred 
re-disentangled latent space.
\begin{equation}\label{eq:lc}
L_c = L_c(c') + L_c(\tilde c').% + L_c(c).
\end{equation}

\section{Experiments}

\subsection{Data}

Consistent with previous works, we use Yelp Service Reviews\footnote{\url{https://github.com/shentianxiao/language-style-transfer}}~\cite{Shen2017Style}, Amazon
Product Reviews\footnote{\url{https://github.com/fuzhenxin/textstyletransferdata}}~\cite{Fu2018Style}, and the GoEmotions dataset\footnote{\url{https://github.com/google-research/google-research/tree/master/goemotions}}~\cite{demszky2020goemotions} as the datasets for explicit disentanglement. In the Yelp dataset, there are $444k$, $63k$, and $126k$ reviews in the train, valid, and test sets, while the Amazon dataset contains $559k$, $2k$, and $2k$, respectively. Both datasets contain sentiment labels with two possible values (``pos'' and ``neg''). Besides, the tense label is also available in the Amazon dataset,  which contains three possible values (``past'', ``now'', and ``future''). 

GoEmotions dataset contains 58,009 examples with the train, test, and validation sets split as 43,410, 5,427, and 5,426 examples respectively. GoEmotions annotations categorize the examples into 27 distinct emotion labels. These emotion labels are further grouped in two ways:
First, by sentiment into positive, negative, and ambiguous classes.
Second, by Ekman's emotion taxonomy which divides the emotions into 6 broad categories: anger (including anger, annoyance, disapproval), disgust, fear (including fear and nervousness), joy (covering all positive emotions), sadness (including sadness, disappointment, embarrassment, grief, and remorse), and surprise (spanning all ambiguous emotions). The mapping relations are shown in Table~\ref{tab:goemo}.

\subsection{Evaluation Metrics}

We borrow the metric mutual information gap (MIG) in \cite{chen2018isolating} for evaluating the disentanglement performance. MIG was originally proposed for implicit disentanglement, which takes each single dimension (a scalar latent variable) of the latent vector as an attribute. In the original design, MIG measures the difference of two mutual information values, one of them is the mutual information between  the
ground truth factor $v_k$ and   latent variable $z_\ast$  ($z_\ast$ is the best fit latent variable for $v_k$ with  the largest mutual information), the other is the mutual information between  the
ground truth factor $v_k$ and   latent variable $z_{\ast\ast}$ ($z_{\ast\ast}$ is the second best fit latent variable for $v_k$). 
MIG is defined as follows %shown  in Eq.~\eqref{eq:mig_implicit}
\cite{chen2018isolating}: 
\begin{align}
% \begin{tiny}
    \text{MIG}_\text{im}&=\frac{1}{K}\sum_{k=1}^K\frac{1}{H(v_k)}\Big(I(z_\ast;v_k)-I(z_{\ast\ast};v_k)\Big), \label{eq:mig_implicit}
 \end{align}   
 where the subscript ``im'' stands for implicit disentanglement, and the mutual information $I(z;v_k)$ is defined by:
 \begin{align}
    I(z;v_k) &= \mathbb E_{q(z,v_k)}\Big[\log\!\sum_{X\in\chi_{v_k}}\!\!\!q(z|X)p(X|v_k)\Big],
    % \end{small}
\end{align}
where $K$ is the  latent vector's dimension, $H(v_k)$ is the entropy of $v_k$, and $\chi_{v_k}$ is the support of $p(X|v_k)$.

When computing MIG in explicit disentanglement, we replace the latent variables $z_\ast$ and $z_{\ast\ast}$ by $s$ and $c$: % as in Eq.~\eqref{eq:mig_explicit}.
\begin{equation}\label{eq:mig_explicit}
    \text{MIG}_\text{ex} = \frac{1}{H(v_k)}\Big(I(s;v_k)-I(c;v_k)\Big),
\end{equation}
 where the subscript ``ex'' stands for explicit disentanglement. 
 
When evaluating the attribute control performance, we have 4 metrics for the NLP tasks.
\begin{table*}[!ht]
\centerline{
	\resizebox{\linewidth}{!}{
\begin{tabular}{lcccc cc| ccccccc}
\toprule[1.0pt]
    & \multicolumn{6}{c|}{Yelp}                            &\multicolumn{7}{c}{Amazon}    \\\cmidrule[0.5pt]{2-14}
 %Methods   & Origin &\multicolumn{3}{c|}{Transfer}  & Origin&\multicolumn{3}{c|}{Transfer}  & Origin&\multicolumn{3}{c|}{Transfer}  \\\cline{2-13}
          & TA & CBLEU-1 & CBLEU-4    &PPL    & TBLEU  & GM  & TA(S) &TA(T) &  CBLEU-1 & CBLEU-4  &   PPL & TBLEU & GM  \\
          \midrule[0.5pt] 
 %\newcite{Fu2018Style}                     & 0.18 & \textbf{0.96} &\textbf{0.67} & 0.40& 0.93& 0.36&    - & - & - \\\hline
 % Heuristic  & 0.726 & 88.5 &58.9 & 213& 76.8&      &   0.545 & - & 85.4& 52.3& 342 & 71.7 &  \\\hline
 \cite{logeswaran2018content}&  0.905&53.0     &7.5     &133  &  17.4  &0.105    & 0.857  & -&  31.5   & 1.8  & 187& 16.6 & 0.091\\
\cite{lample2019multipleattribute}&  0.877    & -    &-     & 48&   14.6&0.139 & 0.896  &-  &-   &-& 92& 18.7&0.122 \\
 \cite{john-etal-2019-disentangled}~(Vanilla)& 0.883 &-&- & 52 & 18.7  &0.147    & 0.720 &-&- &-& 73& 16.5&0.118\\
 \cite{john-etal-2019-disentangled}~(VAE)    & 0.934 &- &-&   32  & 17.9  & 0.174   & 0.822 &-& - &-&  63&9.8&0.109 \\
MTDNA~(Vanilla)      & 0.877 & 30.4 & 4.3 &    45       &   16.1    & 0.146   & 0.789 & 0.963& 23.4 & 1.2& 68& 15.4&0.121 \\
 MTDNA~(VAE)       &  0.944 & 32.6& 5.1  &    27  &   21.2 & 0.195&  0.902&0.993 & 24.0& 1.2&    44& 20.1&0.160 \\
 \cite{qian-etal-2022-controllable}  &   0.873   & -   &  -    &    37  &   - &-&  0.795 &0.902 & -& -&    65& -&-\\
  \midrule[0.5pt]
  \Mname~(Vanilla)   & 0.928  &45.5   &6.9   &  43         &      16.3   &  0.152  & 0.843 & 0.972&  27.6 & 1.5 & 68& 15.9&0.125 \\
  $+$ Back-Translation loss    & 0.890  &54.1  &\textbf{8.7}   &  38         &    16.8    &0.158 & 0.844& 0.975 & 36.7  &2.2  & 59& 17.1 &0.135\\
 $+$ T5   & 0.930  &56.6   &10.4   &  33         &      20.7   &  0.180     & 0.889 & 0.982&  37.4 & 2.4 & 55& 19.3& 0.146 \\
 \Mname~(VAE)       &\textbf{0.951} & 45.7 & 6.3  &    28  &   22.5 &0.197 & \textbf{0.910}& \textbf{0.994 }& 28.2& 1.6&    43& 21.3 &0.165\\
 $+$ Back-Translation loss    & 0.936   & \textbf{54.3}  & 8.4  &   \textbf{26}        & \textbf{22.7}   &\textbf{0.201}    & 0.908& 0.993 & \textbf{37.2}   & \textbf{2.3}  & \textbf{40} &  \textbf{21.7}&\textbf{0.170} \\
  $+$ T5       &\underline{0.985} & \underline{58.1} & \underline{11.2}  &    \underline{25}  &   \underline{23.7} & \underline{0.211 } & \underline{0.921}& \underline{0.994 }& \underline{38.3}& \underline{2.5}&    \underline{38}& \underline{22.9} & \underline{0.177}\\
  \bottomrule[1.0pt]
\end{tabular}
}}

\smallskip 
\caption{Overall attribute control performance.  For the sentiment type, the transfer direction  is  ``Neg$\rightarrow$Pos'', and ``Pos$\rightarrow$Neg''.   For the tense type, the transfer direction  is  ``Past$\rightarrow$Now'', ``Now$\rightarrow$Future'' and ``Future$\rightarrow$Past''. TA(S) is the TA metric for sentiment, while TA(T) is for tense. All the advantages of our results compared to the previous best results are statistically significant, as confirmed by the Wilcoxon signed-rank test. ($p<0.05$). The state-of-the-art results made by pretrained language models are underlined.}
\label{tab:overall}
\end{table*}%

\begin{table*}[!ht]
\centerline{
	\resizebox{\linewidth}{!}{
\begin{tabular}{lcccc ccc}
\toprule[1.0pt]
 %Methods   & Origin &\multicolumn{3}{c|}{Transfer}  & Origin&\multicolumn{3}{c|}{Transfer}  & Origin&\multicolumn{3}{c|}{Transfer}  \\\cline{2-13}
          & TA(Sentiment)& TA(Ekman) & CBLEU-1 & CBLEU-4    &PPL    & TBLEU  & GM-4   \\
          \midrule[0.5pt] 
 %\newcite{Fu2018Style}                     & 0.18 & \textbf{0.96} &\textbf{0.67} & 0.40& 0.93& 0.36&    - & - & - \\\hline
 % Heuristic  & 0.726 & 88.5 &58.9 & 213& 76.8&      &   0.545 & - & 85.4& 52.3& 342 & 71.7 &  \\\hline
 \cite{logeswaran2018content}&  0.723& 0.538& 21.2     &1.5     &224  &  8.9  &   0.111    \\
MTDNA~(Vanilla)      & 0.759& 0.602 & 25.4 & 3.1 &    136       &   9.5    &      0.134 \\
 MTDNA~(VAE)       &  0.780& 0.635& 28.6& 3.7  &    95  &   12.1 &  0.158   \\
 \cite{qian-etal-2022-controllable}  &   0.852& 0.816& -   &  -    &    97  &   - &-\\
  \midrule[0.5pt]
  \Mname~(Vanilla)   & 0.864& 0.845  &34.9   &4.6   &  79         &      15.4   & 0.194     \\
  $+$ Back-Translation loss    & 0.857& 0.832  &36.5  &5.2   &  \textbf{71}         &    17.8    &  0.206 \\
 $+$ T5   & 0.893   & 0.887  &39.7  &7.1   &  63         &      20.3   &  0.225     \\
 \Mname~(VAE)       &\textbf{0.899}&\textbf{0.896} & 36.1 & 5.5  &    76  &   19.4 &0.211 \\
 $+$ Back-Translation loss    &0.886 & 0.858  & \textbf{37.3}  & \textbf{6.6} &  74       & \textbf{21.5}   &    \textbf{0.217}\\
  $+$ T5       &\underline{0.923}& \underline{0.901} & \underline{39.6} & \underline{8.2}  &    \underline{60}  &   \underline{23.3} & \underline{0.238} \\
  \bottomrule[1.0pt]
\end{tabular}
}
}

\smallskip 
\caption{Overall attribute control performance of GoEmotions dataset.  For the sentiment taxonomy, the transfer direction  is  ``Negative$\rightarrow$Positive'', and ``Positive$\rightarrow$ambiguous''.   For the Ekman taxonomy, the transfer direction  is  ``joy$\rightarrow$fear'', ``fear$\rightarrow$sadness'', ``sadness$\rightarrow$disgust'', ``disgust$\rightarrow$anger'', ``anger$\rightarrow$surprise'', ``surprise$\rightarrow$joy''. TA(Sentiment) is the TA metric for sentiment, while TA(Ekman) is for Ekman taxonomy. All the advantages of our results compared to the previous best results are statistically significant, as confirmed by the Wilcoxon signed-rank test. ($p<0.05$). The state-of-the-art results made by pretrained language models are underlined.}
\label{tab:emo_overall}
\end{table*}%

\begin{itemize}

   \item Attribute transfer accuracy (TA): Following previous works~\cite{john-etal-2019-disentangled,sha2021multi}, we use an external sentence classifier (TextCNN~\cite{kim-2014-convolutional}) to measure the sentiment accuracy after the attribute change. The external sentence classifiers are trained separately for the Yelp and the Amazon dataset, and achieved an acceptable accuracy on the validation set (Yelp: $97.68\%$, Amazon: $82.32\%$).
   
    \item Content preservation BLEU (CBLEU-1 \& CBLEU-4): This metric is proposed in \cite{logeswaran2018content}, which transfers the attribute-transferred sentence back to the original attribute, and then computes the BLEU score with the original sentence.
    \item Perplexity (PPL): Perplexity is used for evaluating the fluency of the generated sentences. We use a third-party language model~\cite[KenLM]{kneser1995improved} as the evaluator. Two separate KenLM's are trained and used for evaluation  on the two datasets.
    \item Transfer BLEU (TBLEU): The BLEU score is calculated between the original sentence and the attribute-transferred sentence. We delete the sentiment words before evaluation according to a sentiment word list.\footnote{\url{https://ptrckprry.com/course/ssd/data/positive|negative-words.txt}}
   \item Geometric mean (GM): We use the geometric mean of TA,  1/PPL, and TBLEU as an aggregated score, which considers attribute control performance and fluency simultaneously. 
\end{itemize}

\subsection{Disentanglement Performance}
  
\begin{figure*}[!t]
    \centering
    \begin{tabular}{ccc}
\includegraphics[width=0.33\linewidth]{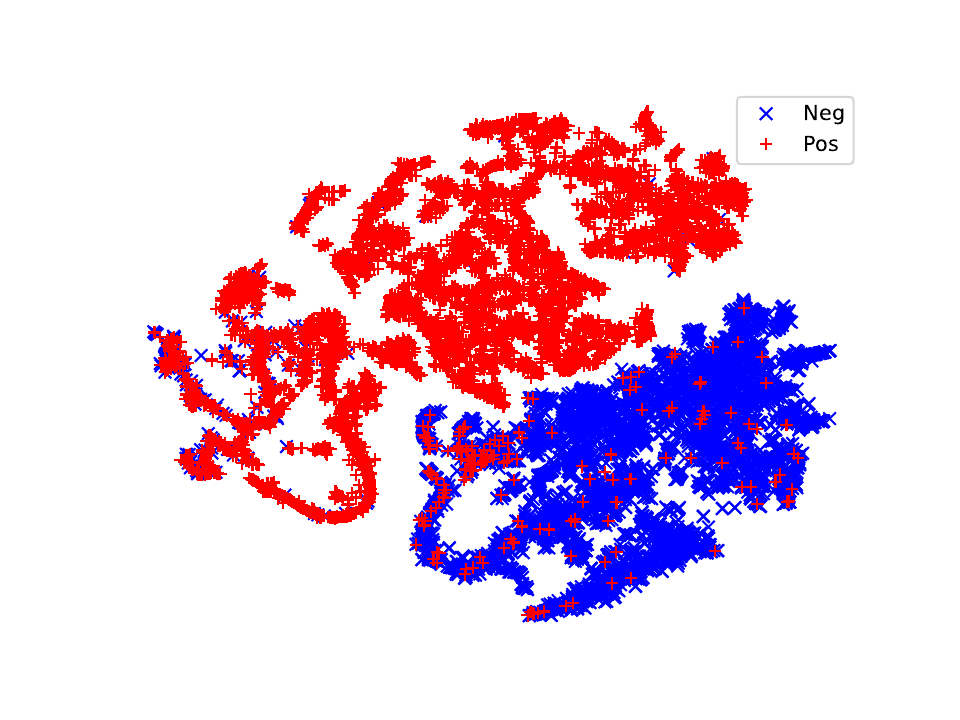}\!\!\!\!\!\! & \includegraphics[width=0.33\linewidth]{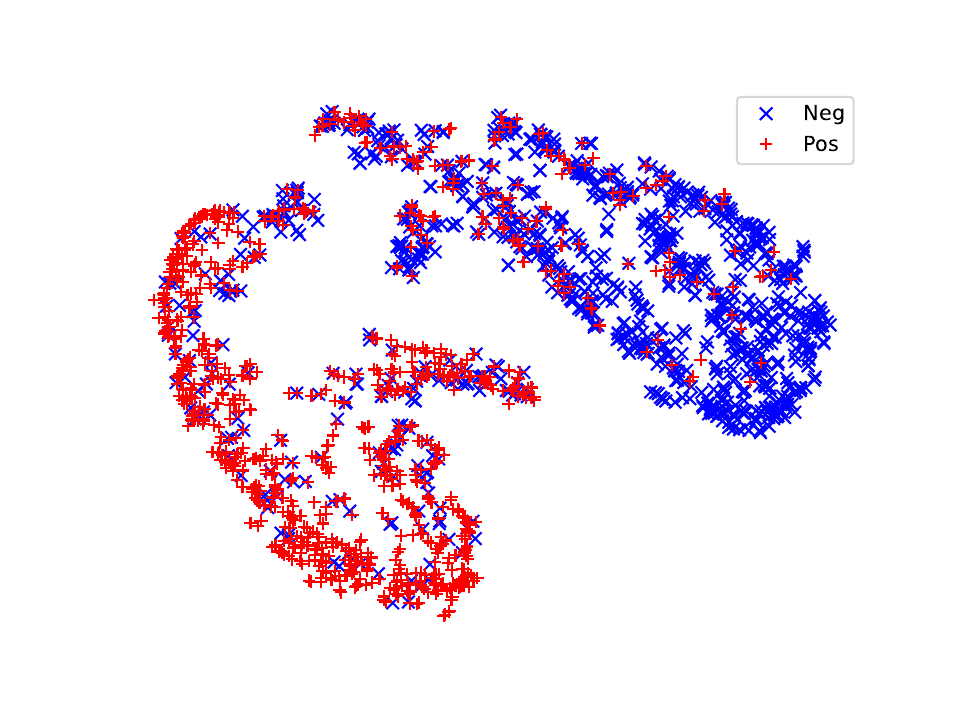}\!\!\!\!\!\! & \includegraphics[width=0.33\linewidth]{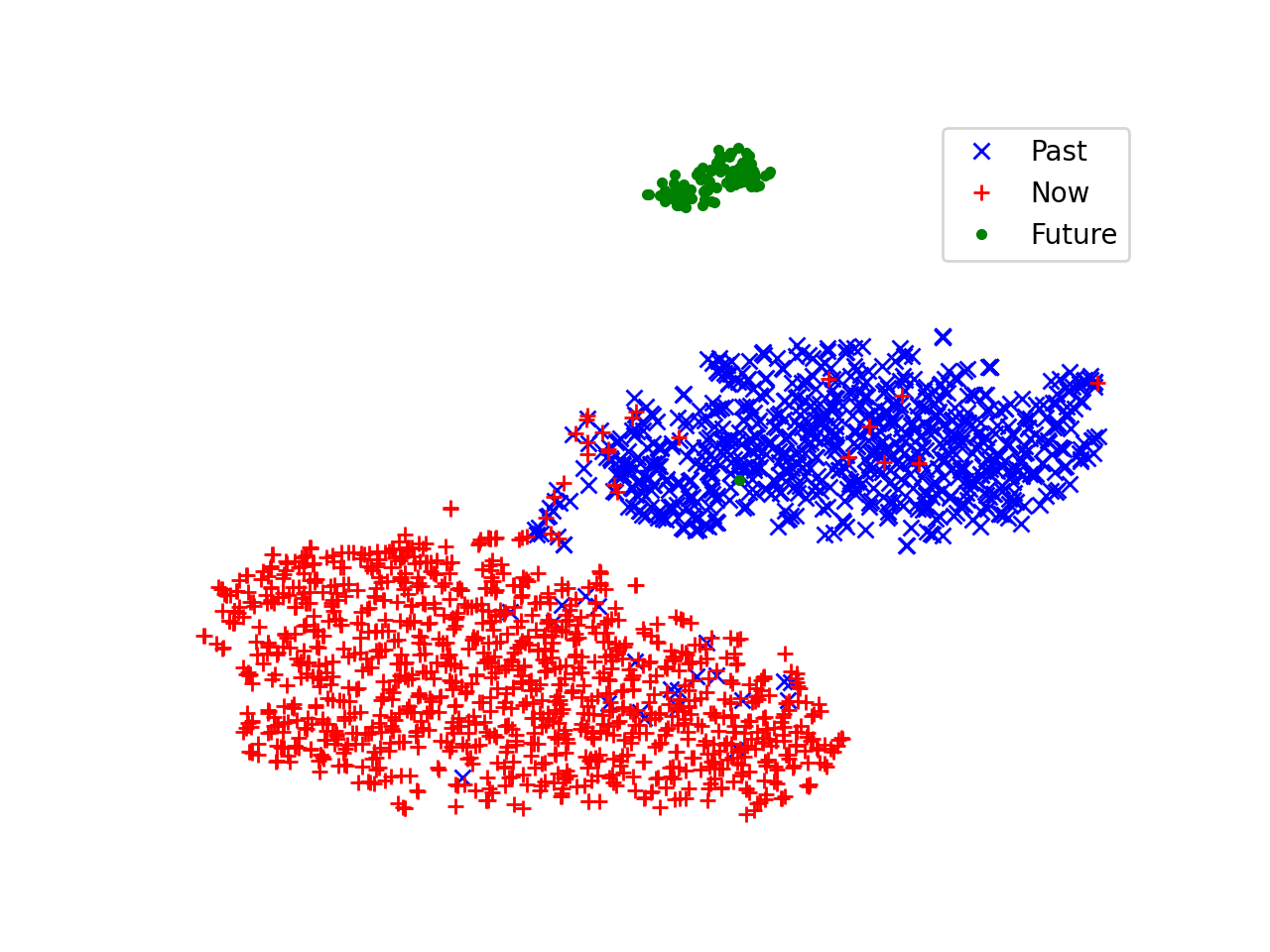}\\
(a) Yelp  (Vanilla) & (b) Amazon  (Sentiment, Vanilla) & (c) Amazon  (Tense, Vanilla) \\
\includegraphics[width=0.33\linewidth]{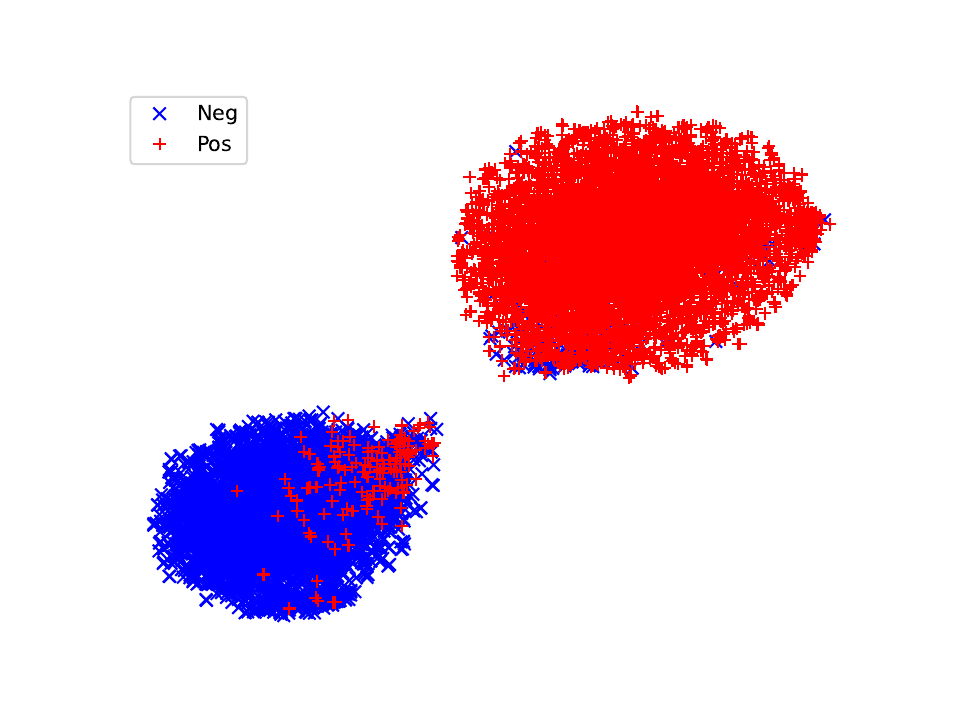}\!\!\!\!\!\! & \includegraphics[width=0.33\linewidth]{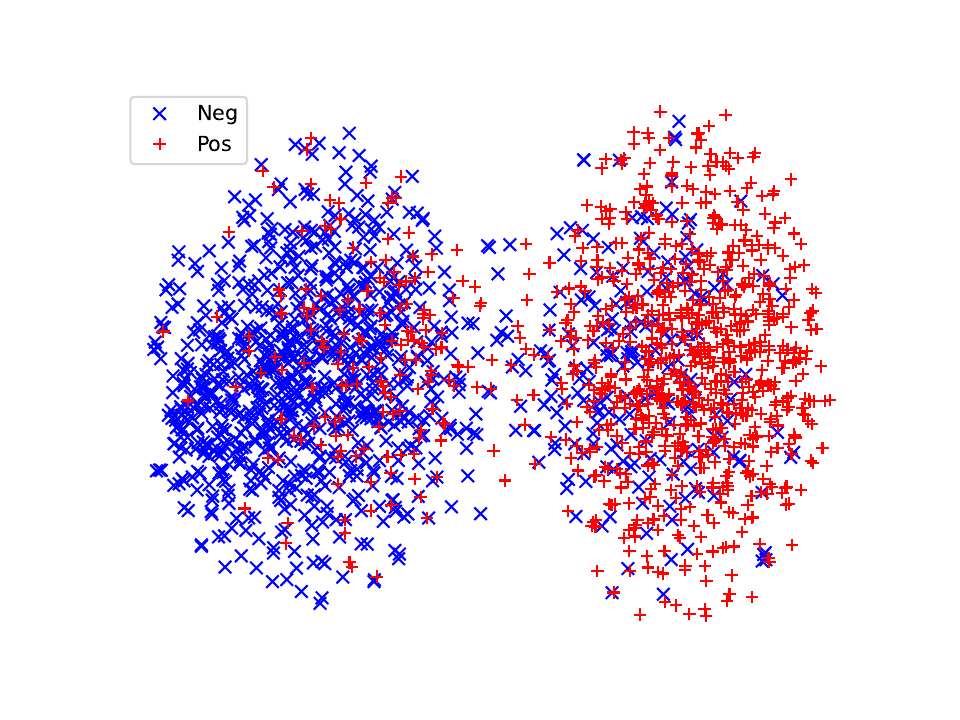}\!\!\!\!\!\! & \includegraphics[width=0.33\linewidth]{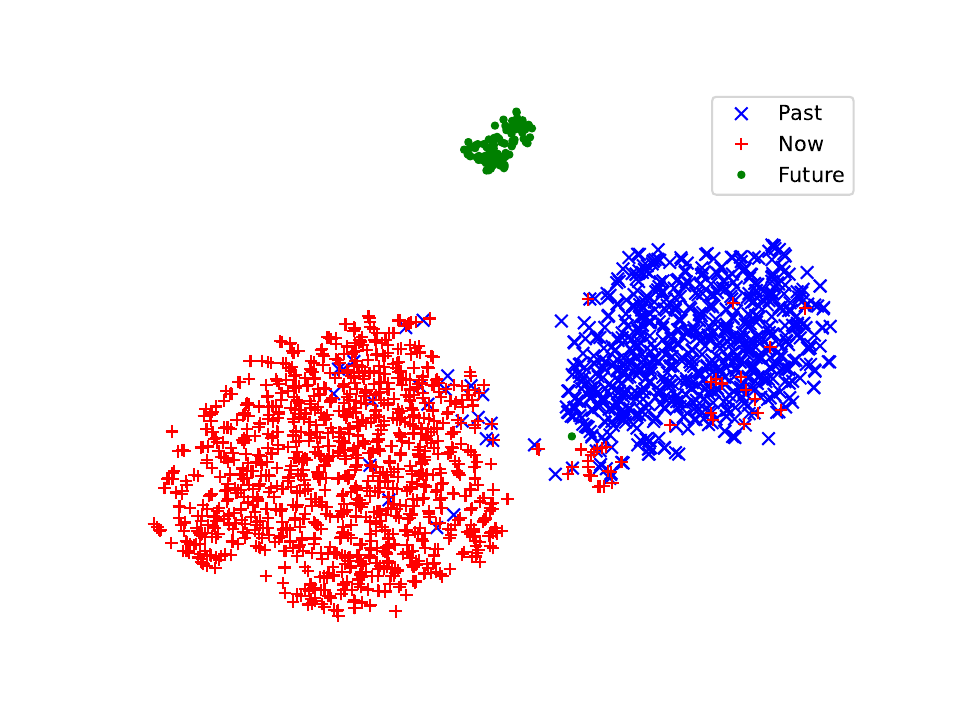}\\
(d) Yelp  (VAE) & (e) Amazon  (Sentiment, VAE) & (f) Amazon  (Tense, VAE)
\end{tabular}
    \caption{Visualization of the disentangled latent space for the two style types: sentiment and tense. (a), (b), and (c) are created by a vanilla autoencoder, while (d), (e), and (f) are created by a VAE. All results are obtained when $\tau$ is set to $100$. }
    \label{fig:vis}%\vspace*{2ex}
\end{figure*}

\begin{figure}
    \centering
    \begin{tabular}{cc}
    \includegraphics[width=0.45\linewidth]{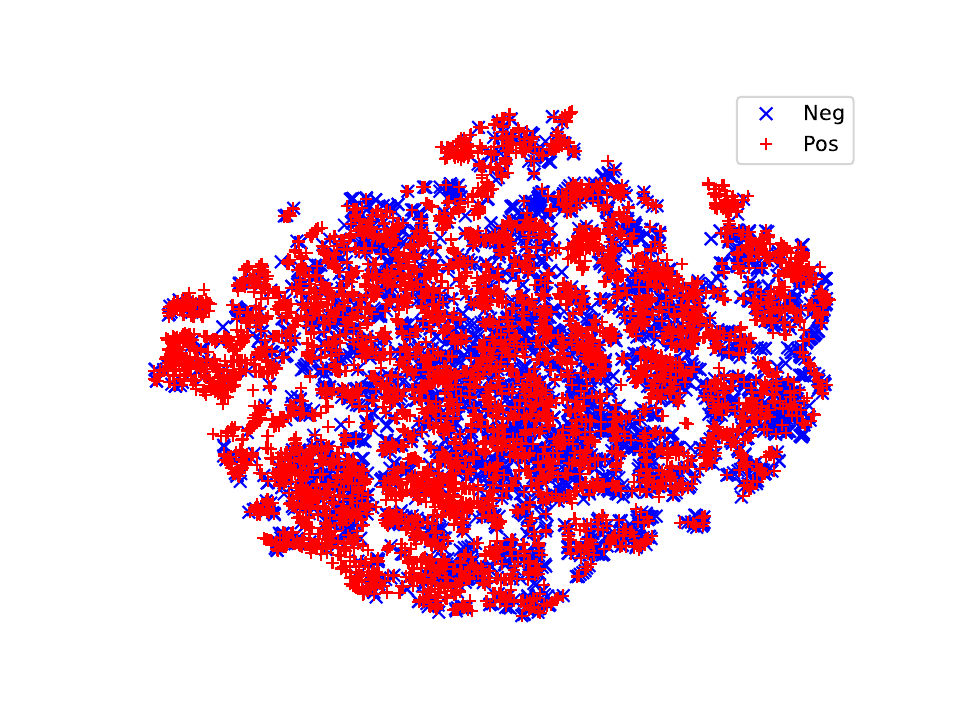}&\includegraphics[width=0.45\linewidth]{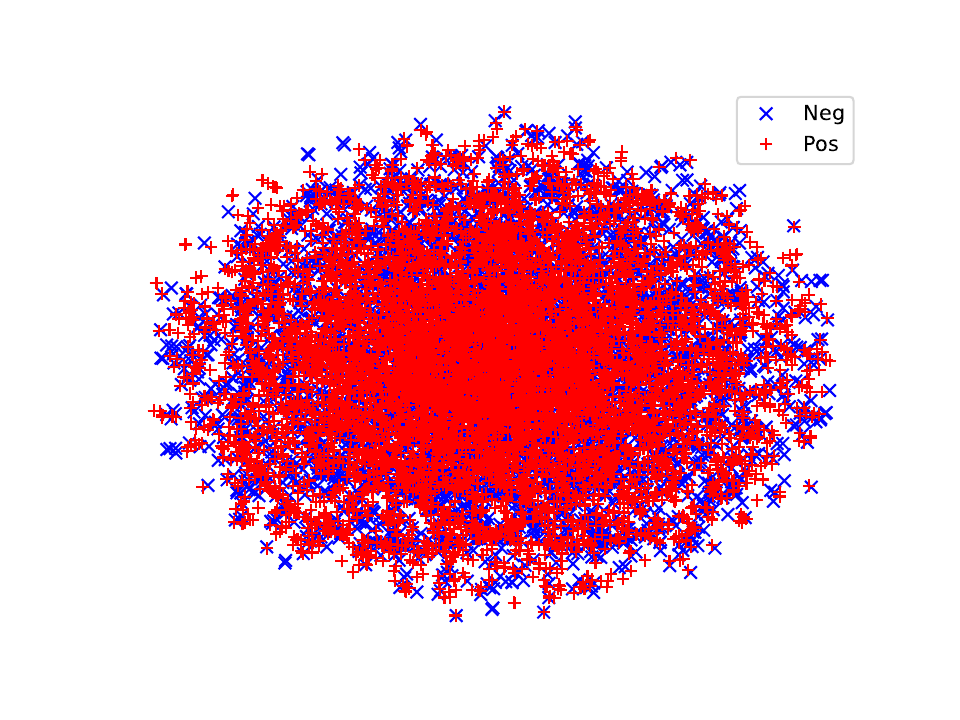}\\
    (a) Vanilla  & (b) VAE\\
    \end{tabular}
    \caption{Visualization of the content spaces after disentanglement on the Yelp dataset.}
    \label{fig:viscontent}
\end{figure}

We have visualized the latent space of attributes and contents in Figures~\ref{fig:vis} and~\ref{fig:viscontent}. To generate this visualization, we perform dimension reduction on the hidden attribute representations in the latent space. Specifically, we use t-SNE~\cite{JMLR:v9:vandermaaten08a} to reduce the high-dimensional attribute representations to 2D embeddings that can be plotted.  We see that with  contrastive learning,  both the vanilla and the variational autoencoder have separated different labels of sentiment (or tense) into different latent spaces successfully.  In comparison, the different labels are mixed together in the content's latent space according to Figure~\ref{fig:viscontent}, which means that the content space does not contain information of the sentiment attribute. Note that we do not use any resource-consuming traditional disentanglement methods like adversarial methods or mutual information minimization, simply re-disentangling the generated sentence and using  contrastive learning can lead to such a good disentanglement performance.

\begin{figure}
    \centering
    \includegraphics[width=0.5\linewidth]{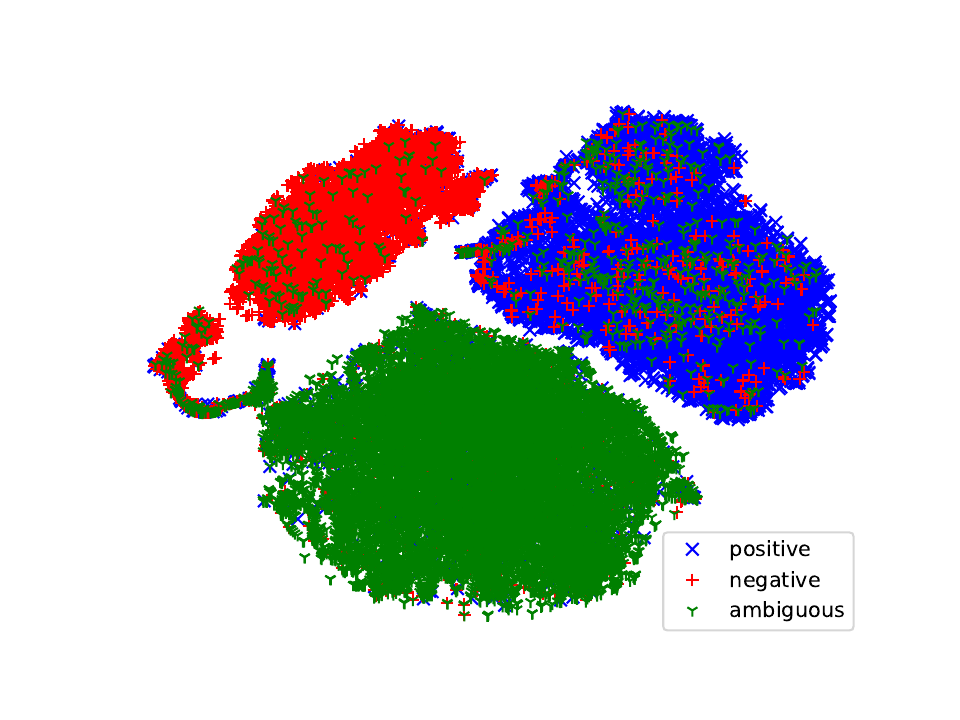}\includegraphics[width=0.5\linewidth]{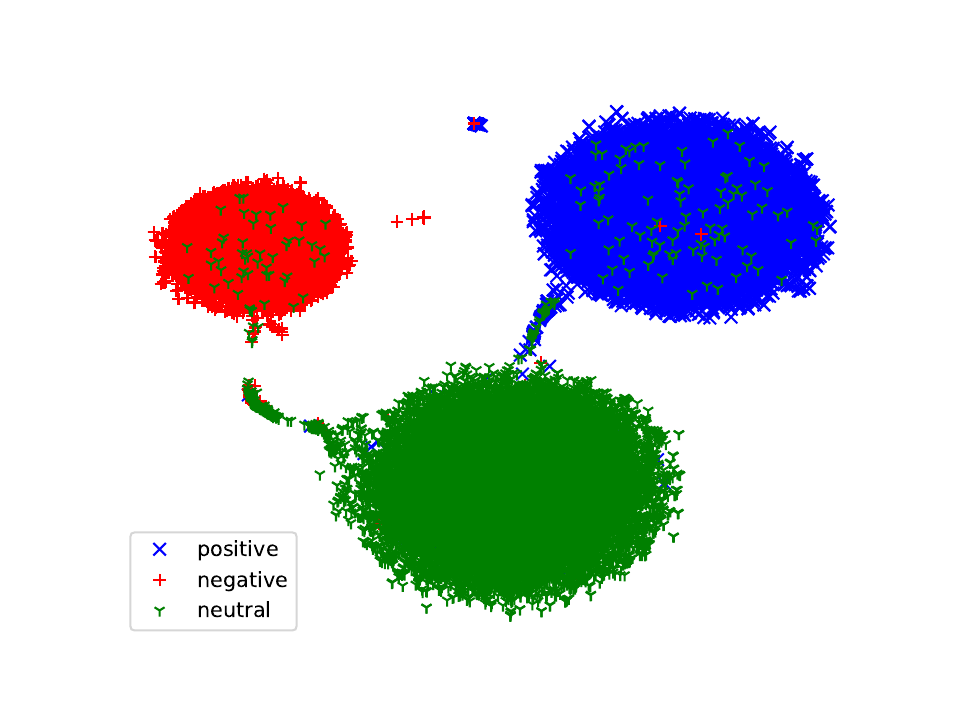}\\
    (a) Sentiment~(Vanilla) ~~~~  (b)  Sentiment~(VAE)\\
    \includegraphics[width=0.5\linewidth]{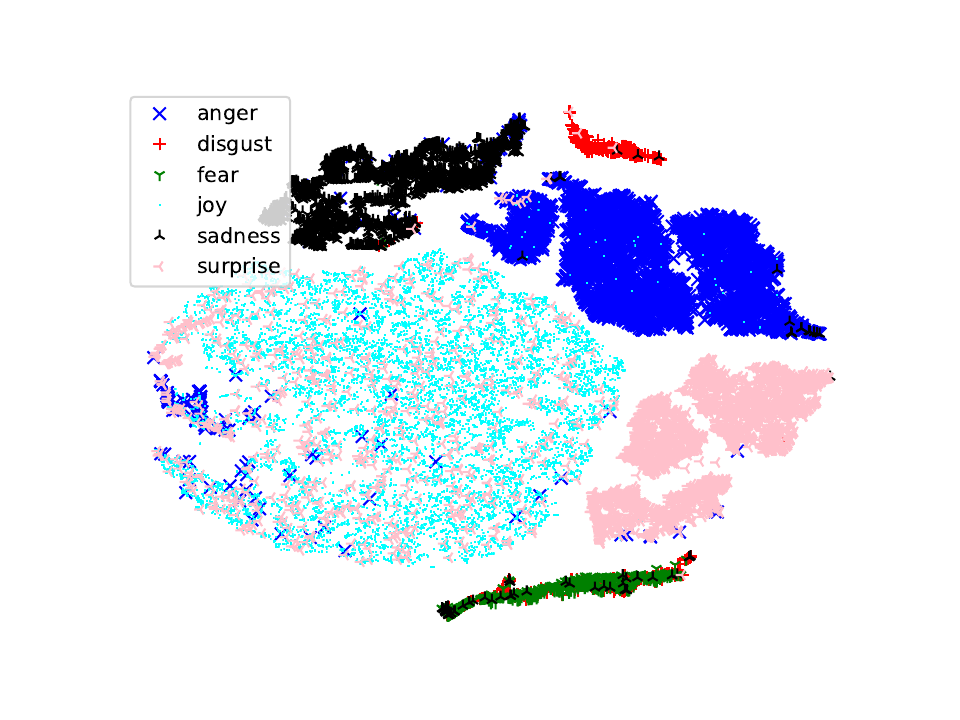}\includegraphics[width=0.5\linewidth]{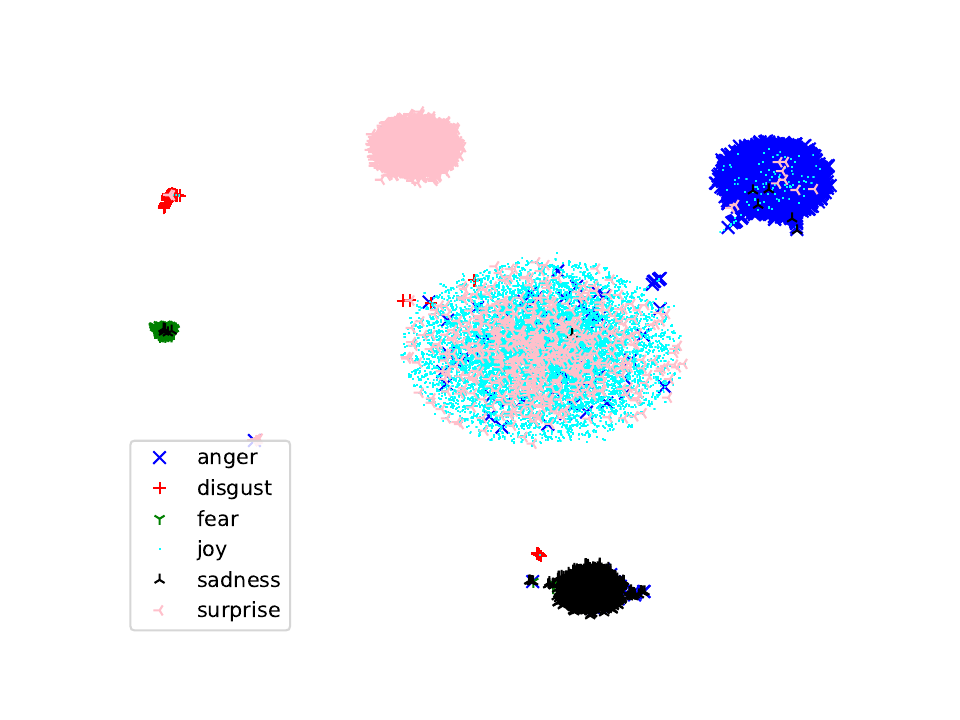}\\
    (c) Ekman~(Vanilla)  \qquad (d)  Ekman~(VAE)\\
    \caption{Visualization of the Sentiment and Ekman taxonomy in the latent space on the GoEmotions dataset. }
    \label{fig:goemo_vis}
\end{figure}

For datasets with more granular emotion categories, we also visualize the attribute latent space of the GoEmotions dataset. We  again use t-SNE to reduce the high-dimensional attribute representations to 2D embeddings that can be plotted.  As shown in Fig.~\ref{fig:goemo_vis}, the 2D latent space naturally separates into three distinct clusters corresponding to the semantic-level taxonomy of positive, negative, and neutral emotions. Furthermore, within the positive and negative regions, the space separates into smaller sub-clusters representing each of the six Ekman emotions. This demonstrates that our model has learned a disentangled latent space where proximity aligns with annotated emotion similarities. By visualizing the latent space in 2D, we can better understand the relationships learned between different emotion categories.

\begin{figure}[!t]
    \centering 
    \includegraphics[width=0.5\linewidth]{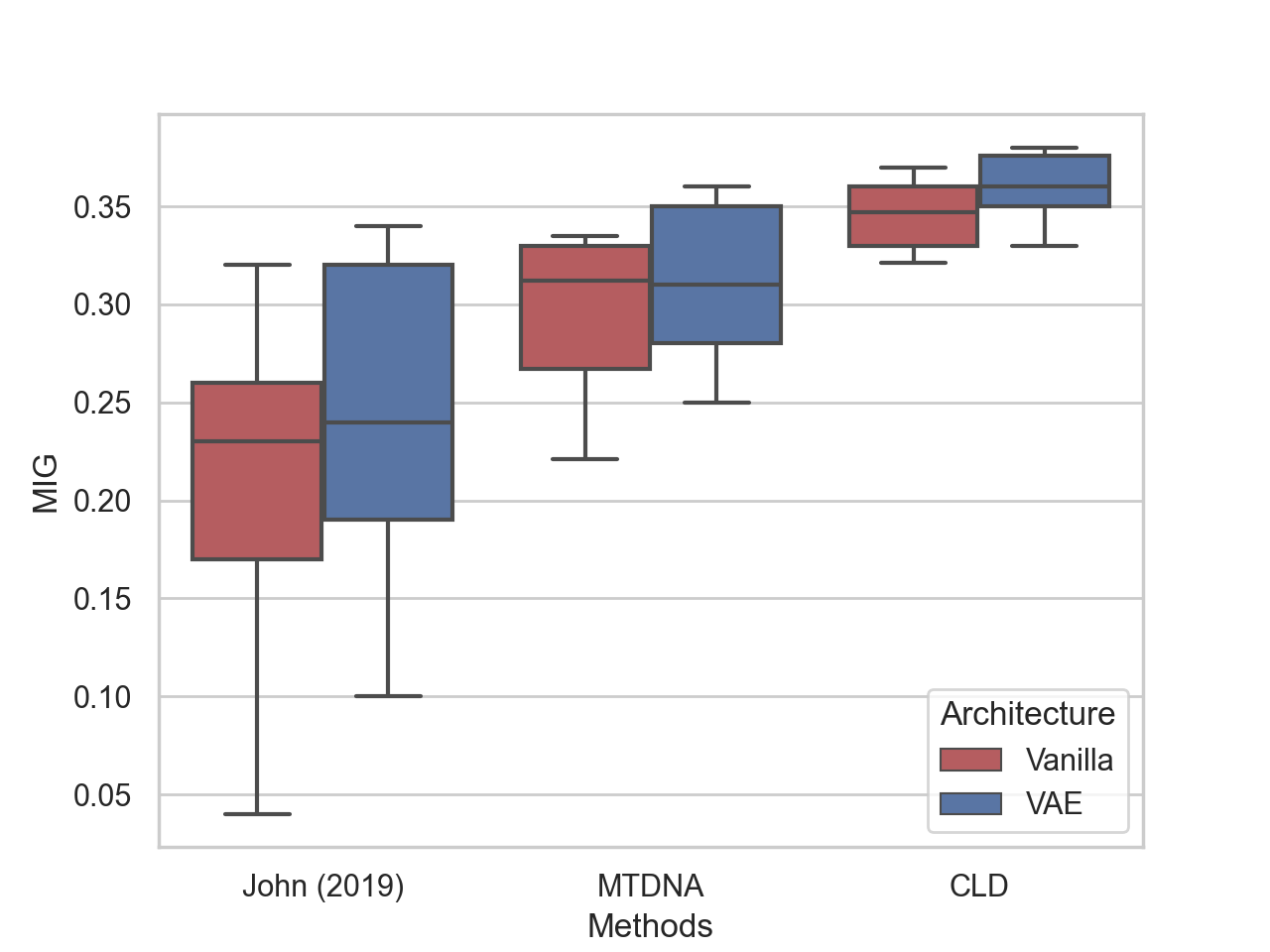}\includegraphics[width=0.5\linewidth]{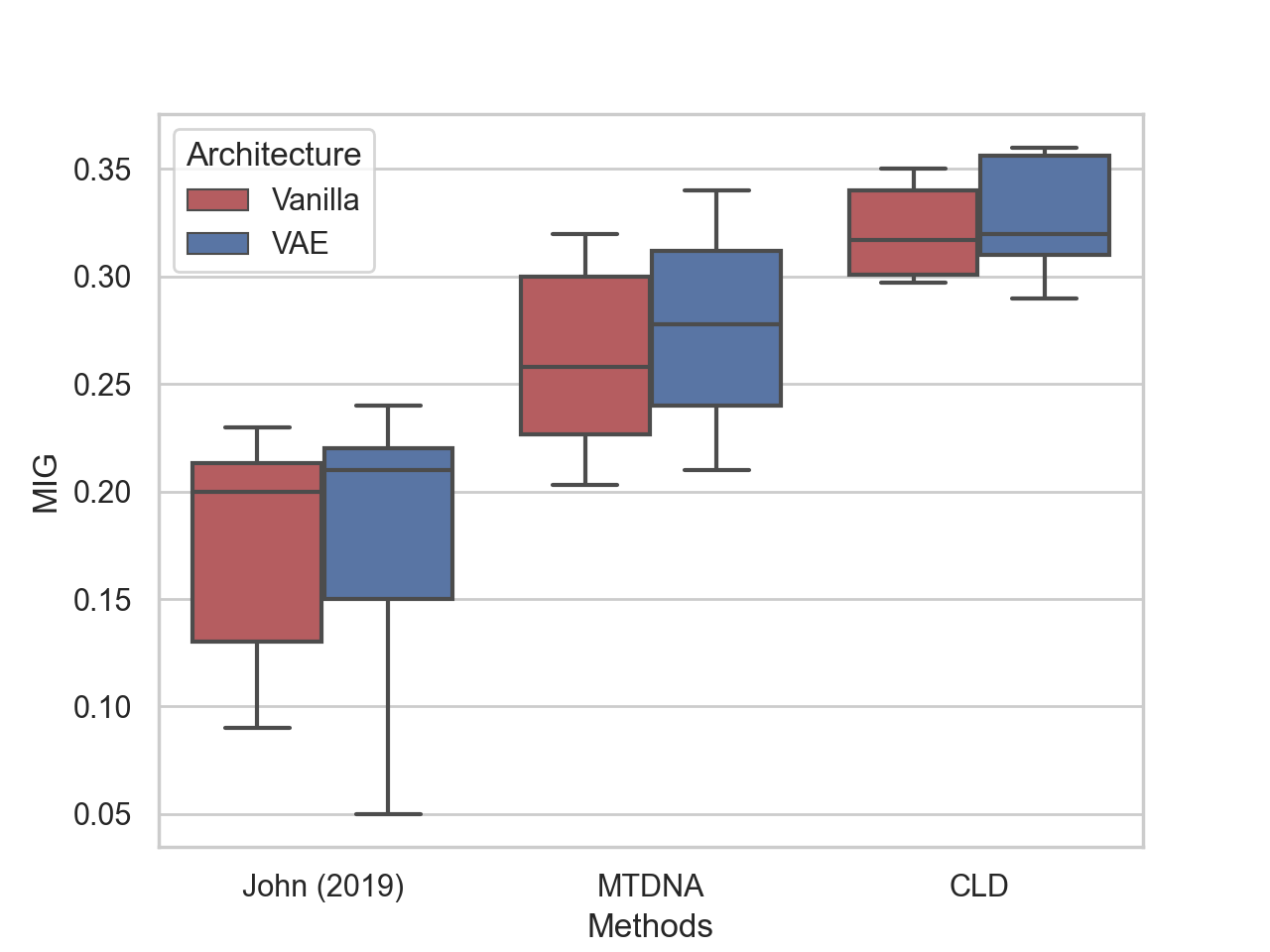}\\
    (a) Yelp dataset \qquad (b) Amazon dataset
    \caption{The box plot of the MIG metric for different explicit disentanglement methods on the two datasets in the experiments (for sentiment). The red boxes represent vanilla-autoencoder-based methods, while the blue boxes are for VAE-based methods.}
    \label{fig:boxex} 
\end{figure}

Also, the comparison of the MIG value is shown in Figure~\ref{fig:boxex}. We reimplemented the previous works of explicit disentanglement, \cite{john-etal-2019-disentangled} and MTDNA~\cite{sha2021multi}, based on their released code, the hyperparameters of the encoder and the decoder are all set to the same. Different experiments for a model would have multiple different MIG values due to different random initialization. So, we draw box plots to show the statistical comparison of MIG values in 40 experiments. In both datasets for explicit disentanglement, our method \Mname\ achieves a better MIG value and is more robust (has smaller variance) than the other two methods.

Besides, due to the computational efficiency of contrastive learning losses, our proposed method takes less time for each epoch  compared to adversarial-based  and mutual-information-based methods.  On Yelp, it takes \Mname\ 20.93 min (Vanilla) and 21.56 min (VAE) for one epoch, while \cite{john-etal-2019-disentangled} requires 46.36 min (Vanilla) and 44.59 min (VAE) for one epoch, MTDNA~\cite{sha2021multi} requires 42.74 min (Vanilla) and 43.62 min (VAE) for one epoch.

\subsection{Performance of Attribute Control}
% \paragraph{Baselines}
We compare our method CLD with multiple previous attribute  control methods:  \cite{logeswaran2018content} and \cite{lample2019multipleattribute} as non-disentanglement methods, and  \cite{john-etal-2019-disentangled}   and MTDNA~\cite{sha2021multi} as explicit disentanglement methods. 
We also compared our approach with the prefix-tuning-based method by \newcite{qian-etal-2022-controllable} for controlling the attribute of generated text. However, we note that their method was not specifically designed to maintain the text content while modifying attributes. Therefore, we limited our comparison to the TA and PPL metrics.

%The sentiment attribute transfer is not a very easy task. We  compared our method with a heuristic approach, which is a very simple method. First, it identifies sentiment-polarity words in the sentence, and then, replaces them with words of the opposite polarity. The identification process is conducted by an MLP that takes each word's embedding as input and outputs the probability of the word being a sentiment-polarity word. The MLP is trained by the sentiment word list provided by \newcite{hu2004mining}. The word with opposite polarity is found using WordNet~\cite{Fellbaum1998wn}.  

The overall performances of the Yelp and Amazon datasets are listed in Table~\ref{tab:overall}.  The overall performance of GoEmotions dataset is listed in Table~\ref{tab:emo_overall}. We can see that our proposed method \Mname\ outperforms all the previous works in the transfer accuracy metric (TA), perplexity, and TBLEU score. Compared with the baseline methods without contrastive learning, our approach shows great advantages over the MTDNA~\cite{sha2021multi} models in the CBLEU metrics. This fact shows that the content of a sentence is much easier to be preserved (the attribute control process is more robust) when we are using contrastive learning  to keep the content vector before and after re-disentanglement to be as close as possible.  Moreover, when we added back-translation loss as is conducted by \newcite{logeswaran2018content} and \newcite{lample2019multipleattribute}, our method \Mname\ achieved an even higher score in the CBLEU-1  and CBLEU-4 metric,  and this score has outperformed the state-of-the-art CBLEU score. This again proved that back-translation loss  will become more powerful in content preservation when used together with contrastive learning. According to the aggregated performance (GM) listed in Table~\ref{tab:overall}, \Mname\ also outperforms the baseline methods, and \Mname(VAE) with  back-translation loss achieved state-of-the-art results. We have observed similar results in the tense attribute, which is shown in the column ``TA(T)'' in Table~\ref{tab:overall}.

We also conducted a comparison between our method and the prompt-tuning-based approach proposed by \newcite{qian-etal-2022-controllable}. However, it is important to note that the prompt-tuning-based method only focuses on controlling the attribute of the generated text, without ensuring content preservation. Therefore, we limited our comparison to the TA and PPL metrics. To evaluate their work, we applied \newcite{qian-etal-2022-controllable}'s method on our datasets and assessed the results based on our metrics. As demonstrated in Table~\ref{tab:overall}, our method still has a clear advantage over the prompt-tuning-based approach, as the latter sacrifices some attribute accuracy in order to achieve controllable text generation.

Our method is very easy to be merged with pretrained  language models in encoder-decoder architectures (like T5~\cite{raffel2020exploring}). We merged our  method with T5 and report the results in Table~\ref{tab:overall}. Due to the large storage of text corpus and common sense knowledge in the pretrained language model, the result achieved a much better level in style transfer accuracy, content preservation, and fluency metrics.
%\paragraph{performance desc}
%\paragraph{analysis}

\subsection{Ablation Test}
\begin{figure}[!t]
\centering
\includegraphics[width=0.5\linewidth]{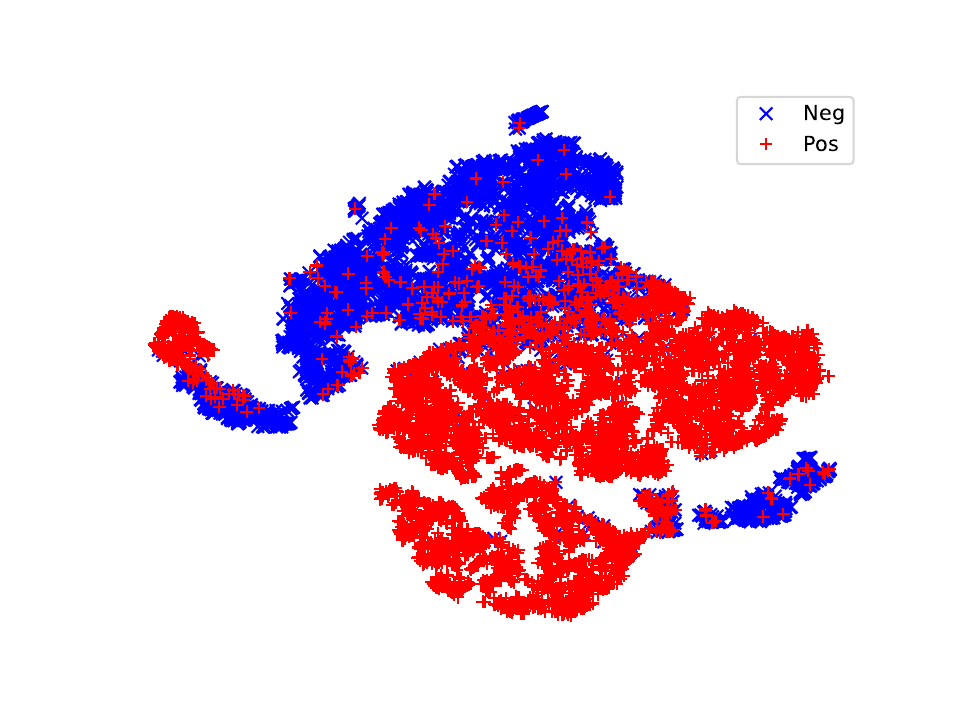}\!\!\!\!\!\!\includegraphics[width=0.5\linewidth]{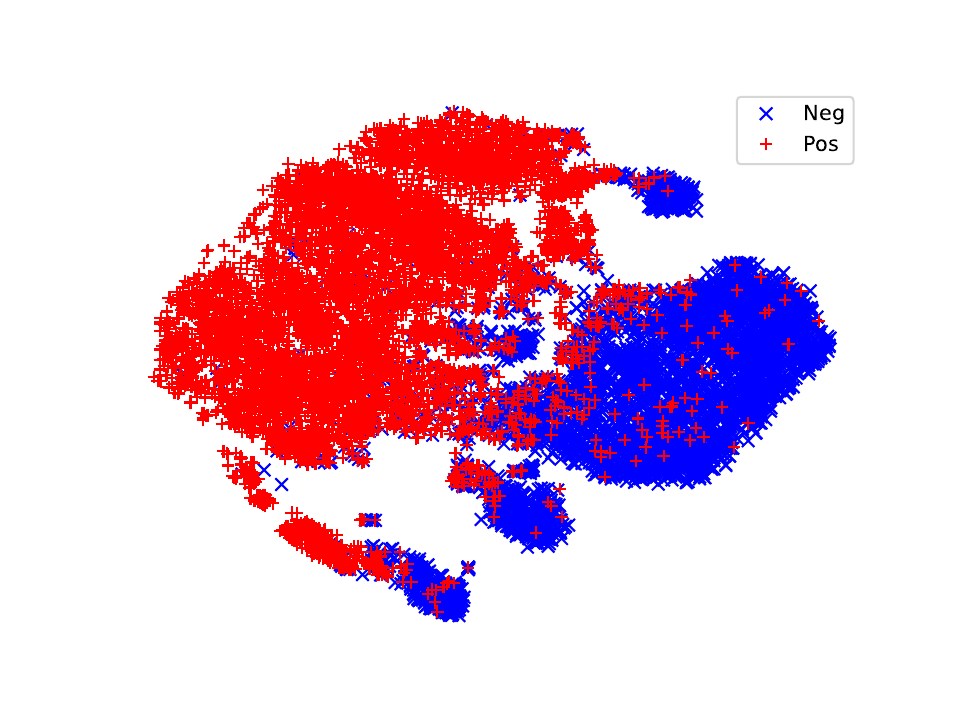}\!\!\!\!\!\!\\
(a) Yelp (Vanilla) \qquad (b) Yelp (VAE)\\
\caption{Visualization of latent space when we remove the re-disentanglement  process, i.e., we only keep the contrastive losses in Eqns.~\ref{eq:7}. }
\label{tab:7_10}
\end{figure}

\paragraph{Effect of Re-disentanglement Process.}
To prove that the re-disentanglement process is necessary, we remove all the contrastive losses related to the re-disentanglement process. The visualization of the latent spaces for vanilla and VAE are shown on Figure~\ref{tab:7_10}. We can see that the latent space  became partly mixed up, which shows that the re-disentanglement process is indispensable. 

\paragraph{Effect of Contrastive Loss Functions.}
To study the effect of each contrastive learning loss, we remove the loss functions one by one to check the difference of the evaluation metrics. The results are shown in Table~\ref{tab:ablation}. We found that after the content contrastive loss $L_c$ is removed, the style transfer accuracy has been improved, which shows that the constraint on the content vector would negatively affect the style information in the generated sentences. Also, the CBLEU-4 and TBLEU scores largely dropped, which shows that $L_c$ is very important for content preservation. Then, after $\tilde L_k$ is removed,  the TA metric dropped about $3$ percentage points, while the CBLEU-4 and TBLEU scores did not have any significant change. Since $\tilde L_k$ is a constraint for the re-disentangled style vector of the style-transferred sentence, it does not have too much  effect on the content of the sentence. A similar phenomenon is observed when we remove the loss $L_\text{re}$: the TA metric significantly decreased again, and the  BLEU scores slightly decreased.

Besides, we also remove the three contrastive learning losses for the content preservation ($L_c(c')$, $L_c(\tilde c')$) to study their effect on the results. The scores are also listed in Table~\ref{tab:ablation2}. We can see that removing any one of the two losses would cause an increase in the TA score, which means all of the content preservation losses are limitations on the style latent space. Both the CBLEU-4 and TBLEU scores decrease a lot after removing the two content preservation losses. In particular,  it seems that $L_c(c')$ has the largest effect on the scores, which is sensible, because a more distinguishable content space is easier for content preservation  intuitively.

\begin{table}[!t]
\centering
\resizebox{\linewidth}{!}{
\begin{tabular}{lcc c}
\toprule[1.0pt]
    & \multicolumn{3}{c}{Yelp}                            \\\cmidrule[0.5pt]{2-4}
          & TA & CBLEU-4      & TBLEU     \\
          \midrule[0.5pt]
  \Mname~(Vanilla)   &  0.928   &\textbf{6.9}       &      16.3       \\
  \Mname~(Vanilla) - $L_c$  & 0.935   & 4.6         &     11.5       \\
   \Mname~(Vanilla) - $L_c$ -$\tilde L_k$   & 0.903   & 4.3             &    10.8       \\
   \Mname~(Vanilla) - $L_c$ -$\tilde L_k$-$L_\text{re}$  & 0.862 & 4.4  &       10.2                  \\
 \Mname~(VAE)       &0.951  & 6.3   &   \textbf{22.5}    \\
   \Mname~(VAE) - $L_c$  & \textbf{0.959}    & 4.2        &   13.6         \\
   \Mname~(VAE) - $L_c$ -$\tilde L_k$    & 0.928 & 4.3         &  12.8          \\
   \Mname~(VAE) - $L_c$ -$\tilde L_k$-$L_\text{re}$  & 0.887 & 4.1                &    12.4        \\
   \midrule[0.5pt]
   \Mname~(Vanilla)~(MSE)   & 0.926   &5.0         &      12.2       \\
    \Mname~(VAE)~(MSE)       & 0.945  & 5.1    & 15.6  \\
   \bottomrule[1.0pt]
\end{tabular}
}
\caption{Ablation test results. We select three metrics (TA, CBLEU-4, and TBLEU) in this experiment, because  they are closely related to the contrastive losses $L_\text{re}$, $\tilde L_k$, and $L_c$. }%\vspace*{-2ex}
\label{tab:ablation}  
\end{table}
   
   \begin{table}[!t]
\centering
\resizebox{\linewidth}{!}{
\begin{tabular}{lcc c}
\toprule[1.0pt]
    & \multicolumn{3}{c}{Yelp}                            \\\cmidrule[0.5pt]{2-4}
          & TA & CBLEU-4      & TBLEU     \\
          \midrule[0.5pt]
   \Mname~(Vanilla) - $L_c(c')$  & 0.929   & 5.2         &    14.8       \\
   \Mname~(Vanilla) - $L_c(\tilde c')$  & 0.930   & 6.1         &     15.3       \\
   %\Mname~(Vanilla) - $L_c(c)$  & 0.932   & 4.9         &     13.7       \\
 \Mname~(VAE) - $L_c(c')$      &0.955  & 5.1   &   17.8    \\
 \Mname~(VAE) - $L_c(\tilde c')$      &0.951  & 5.8   &   20.1    \\
% \Mname~(VAE) - $L_c(c)$      &0.956  & 5.0   &   14.9    \\
 \bottomrule[1.0pt]
\end{tabular}
}
\caption{Ablation test results w.r.t. different components in  $L_c$. }%\vspace*{-2ex}
\label{tab:ablation2}  
\end{table}

\begin{figure*}
    \centering
   
    \begin{tabular}{cccc}
     \includegraphics[trim=40 40 40 40,clip,width=0.23\textwidth]{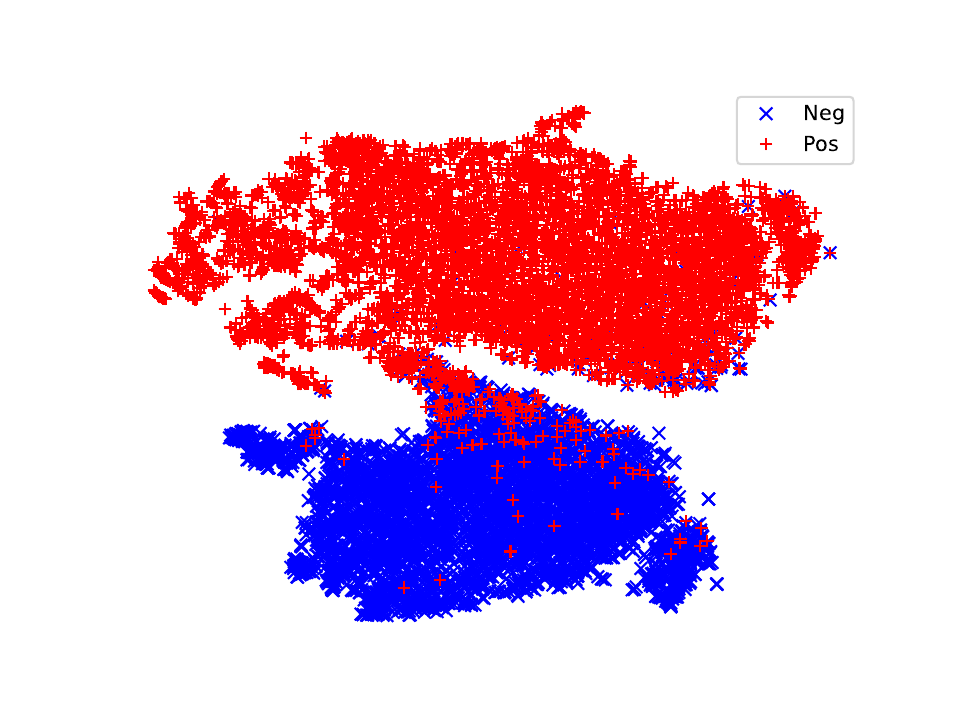}&
     \includegraphics[trim=40 40 40 40,clip,width=0.23\textwidth]{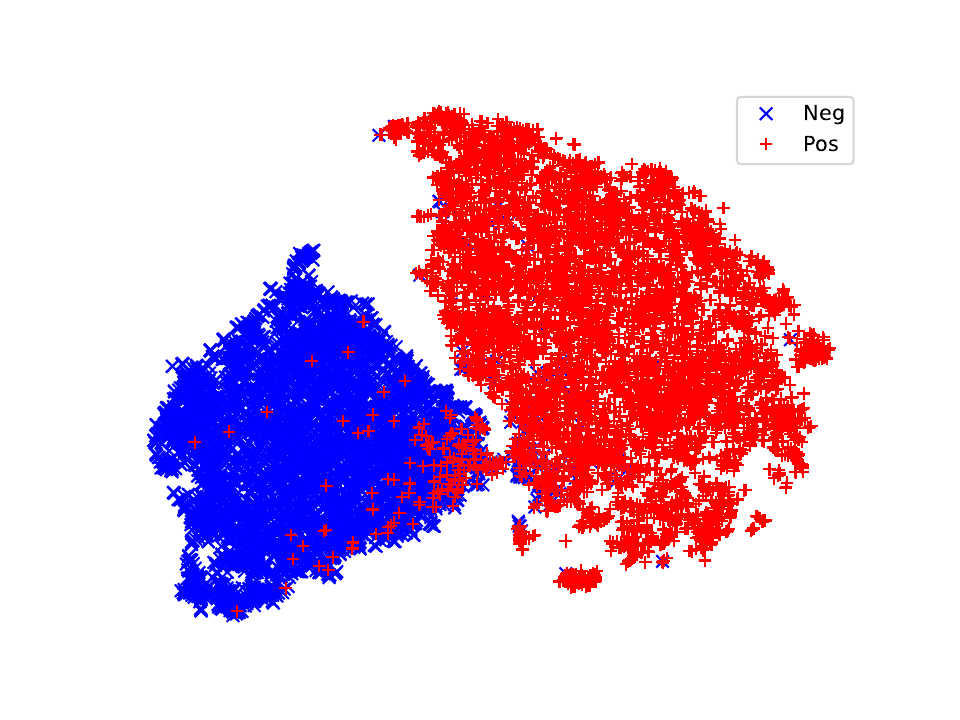}&
     \includegraphics[trim=40 40 40 40,clip,width=0.23\textwidth]{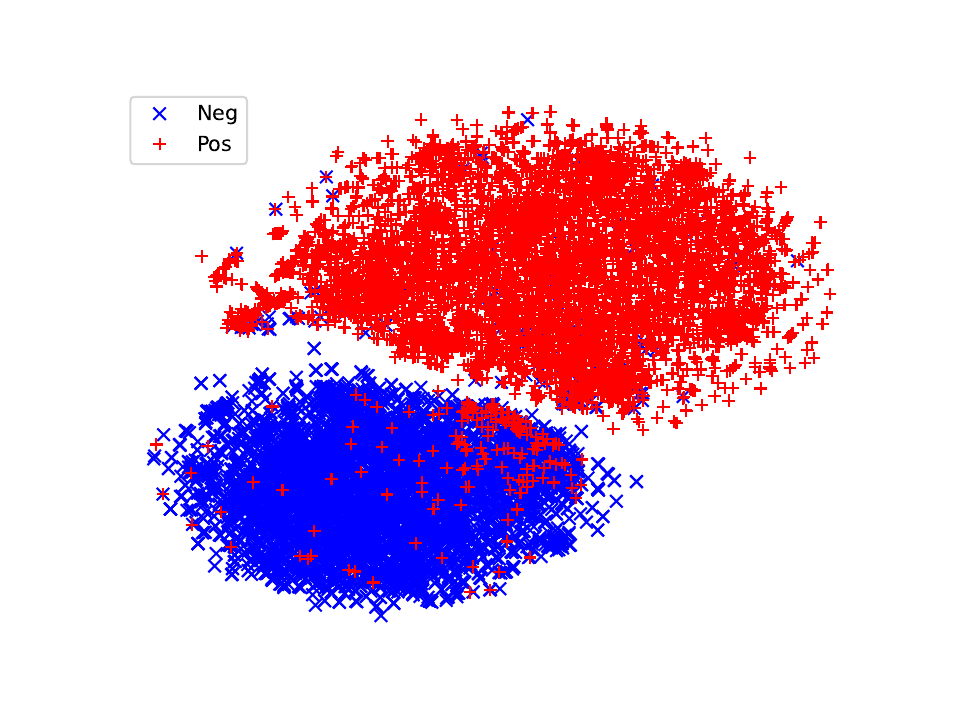}&
     \includegraphics[trim=40 40 40 40,clip,width=0.23\textwidth]{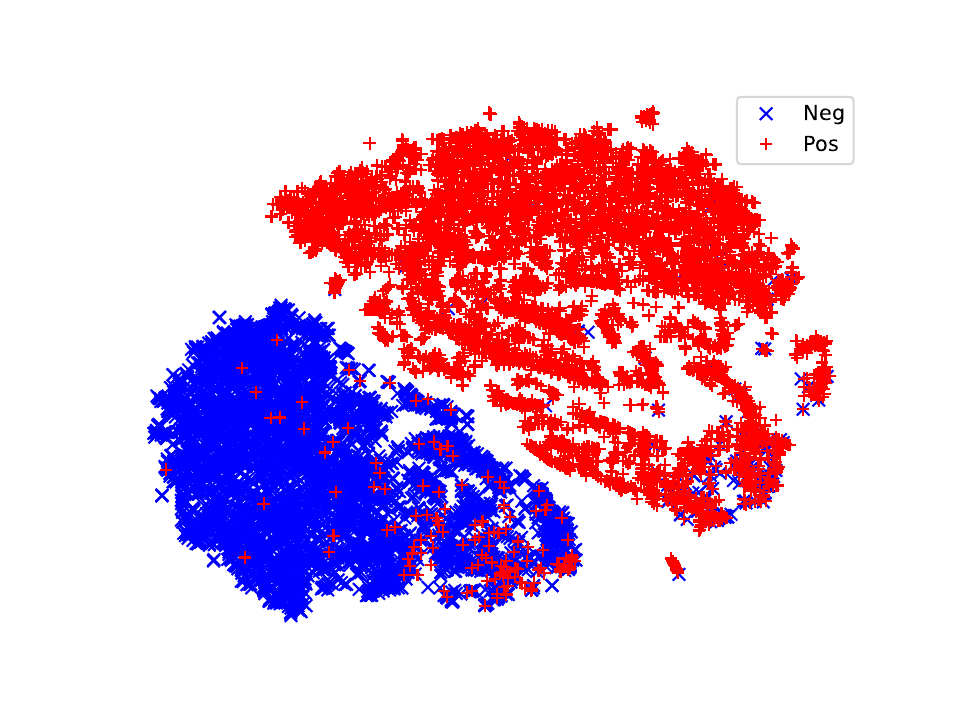}\\
     $\tau=0.5$ & $\tau=1.0$ & $\tau=10.0$ & $\tau=100.0$ \\
     \includegraphics[trim=40 40 40 40,clip,width=0.23\textwidth]{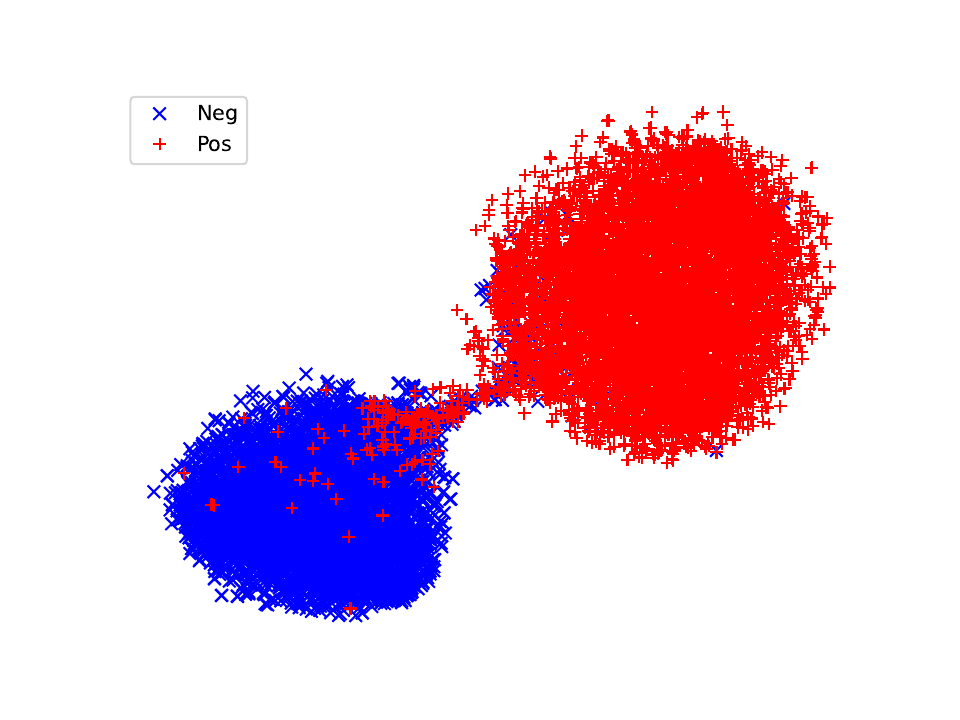} &
      \includegraphics[trim=40 40 40 40,clip,width=0.23\textwidth]{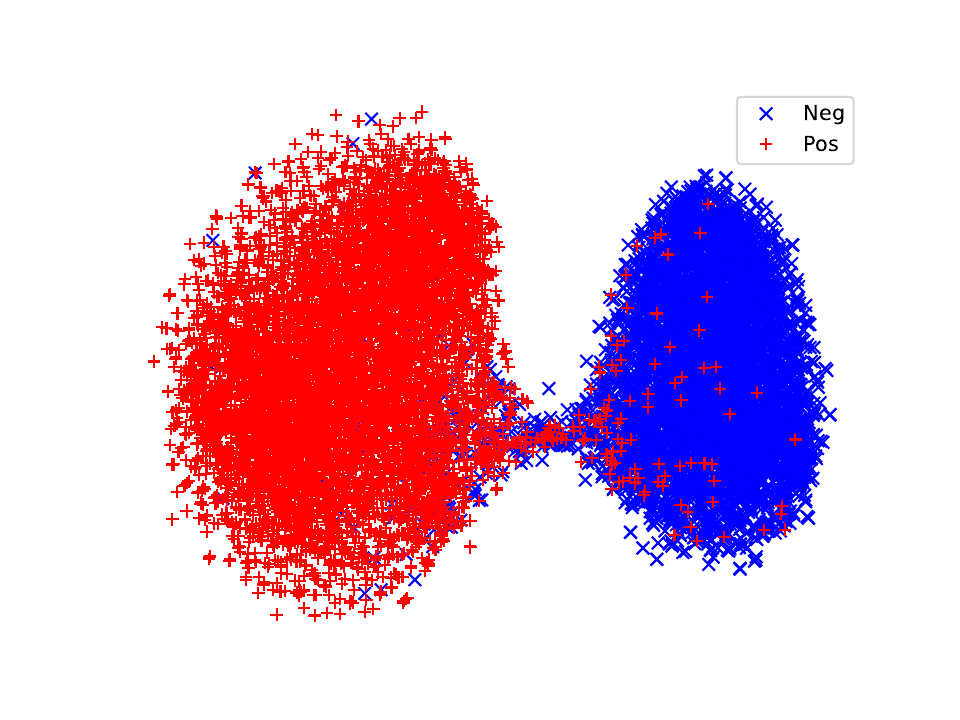} &
      \includegraphics[trim=40 40 40 40,clip,width=0.23\textwidth]{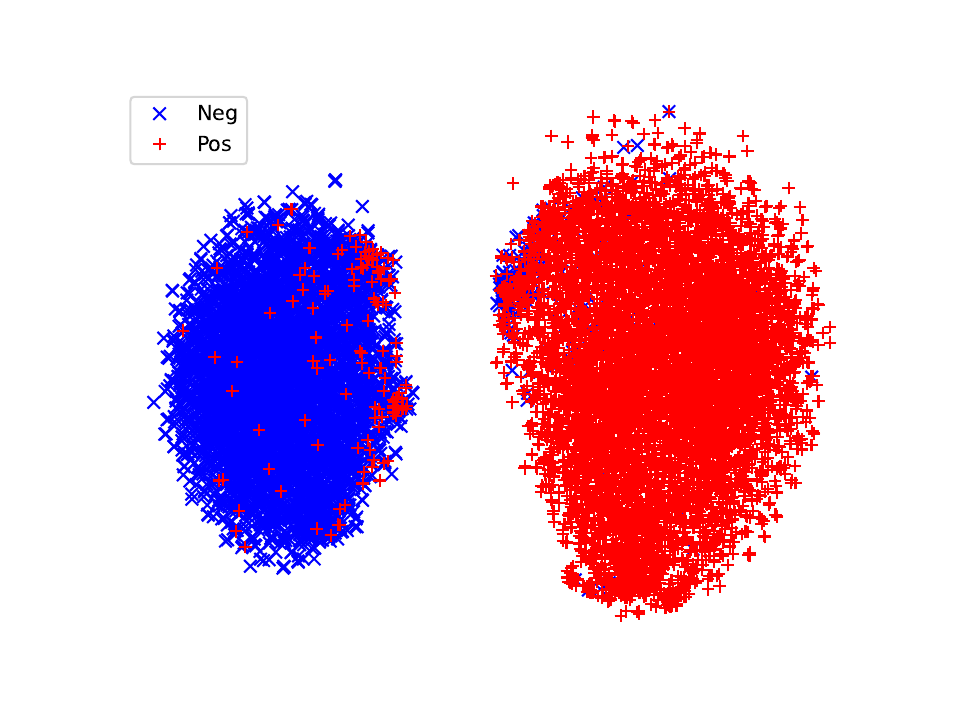} &
      \includegraphics[trim=40 40 40 40,clip,width=0.23\textwidth]{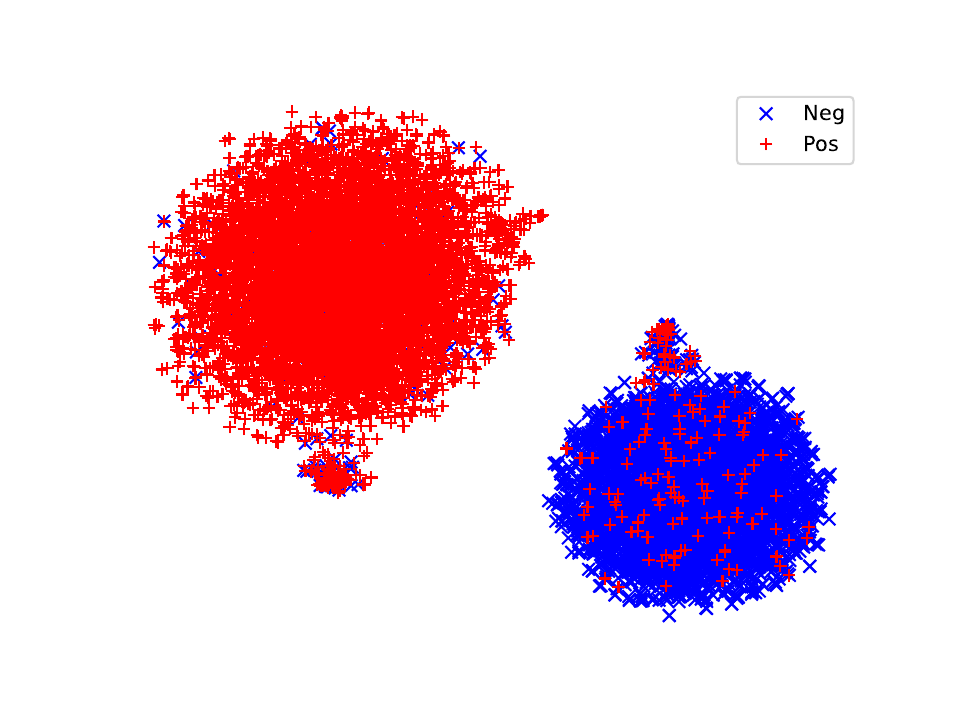}\\
     $\tau=0.5$ & $\tau=1.0$ & $\tau=10.0$ & $\tau=100.0$ \\
    \end{tabular}
    \caption{Change of the latent space when the temperature hyperparameter $\tau$ is getting larger. We show four different $\tau$ values (namely,  ${0.5, 1.0, 10.0, 100.0}$) for the two possible architectures. The first row is from the vanilla autoencoder architecture, while the second row is from the VAE architecture.}
    \label{fig:tau}
\end{figure*}
We also conducted experiments about changing the content's contrastive learning loss $L_c$ to mean-square error (MSE) loss to check whether contrastive learning is necessary. In this experiment, we replace $L_c$ with the following loss $L_\text{mse}$:
\begin{equation}
    L_\text{mse} = \|c' - c\|_2\,,
\end{equation}
where $\|\cdot\|_2$ represents the 2-norm.
The results are also shown in  line \Mname~(Vanilla)~(MSE) and \Mname~(VAE)~(MSE) of Table~\ref{tab:ablation}. We can see that, the score of CBLEU-4 and TBLEU dropped a lot compared to \Mname~(Vanilla) and \Mname~(VAE) after we replaced $L_c$ with   $L_\text{mse}$. The intrinsic difference between $L_c$ and   $L_\text{mse}$ is that $L_\text{mse}$ only encourages $c'$ and $c$ from the same case to be close, while $L_c$ also requires the content vectors from different cases to be far away form each other. The latter alleviates the possibility of the content space to  collapse. This result proved that the contrastive learning loss is inevitable for content preservation.

\paragraph{Effect of $\tau$.}
To investigate the effect of the temperature hyperparameter $\tau$, we run the model several times with different values of $\tau$, and visualize the latent space in Figure~\ref{fig:tau}. According to Figure~\ref{fig:tau}, when $\tau$ has a small value, the latent spaces for the different style values tend to be connected in some area. In contrast, the latent spaces get separated when the value of $\tau$ increases. The reason is that when the temperature  $\tau$ is getting large, the distinction between the positive and negative examples in the contrastive losses tends to be underestimated. Hence, the model needs to work harder to make the distinction large, and thus the latent spaces are getting more separated.

\subsection{Case Study} \label{sec:casestudy}
We sampled some generated text when we are transferring the sentiment attribute from one to another, the results  are shown in Table~\ref{tab:morecasemg}. According to the results, the content of text almost remains unchanged,  while the target attribute was changed to what we expected. 

Furthermore, we evaluated more complex emotion attribute transfer cases from the GoEmotions dataset. We transformed the emotions according to the Ekman taxonomy and presented the results produced by \Mname\ using both the vanilla and VAE architectures. These results are tabulated in Table~\ref{tab:morecasetense_goemo}.
\begin{table*}[!t]
\begin{center}
\resizebox{\linewidth}{!}{
\begin{tabular}{p{6cm}p{6cm}p{6cm}}
\toprule[1.0pt]
\textbf{Original (Pos)} & \textbf{Vanilla Transferred (Neg)} & \textbf{VAE Transferred (Neg)}\\
%i will not be purchasing more of these as there are other toys that they like better . &i was not expecting a toy of that at all about that to be as good as . &i tried my other ones as a gift to someone who loves to play UNK the UNK .\\
\midrule[0.1pt]
every one is so nice , and the food is amazing ! & the servant is rude and the food is terrible .  & every one is so tepid , and the food is awful. \\
\midrule[0.1pt]
an excellent dining experience . & the dining feels bad . &an awful dining experience . \\
\midrule[0.1pt]
yesterday i went to this location and the staff was very informative and personable . & yesterday i went to this location and found the staff so rude and angry . & yesterday i went here and the staff was very tepid , not a good choice  . \\
\midrule[1.0pt]
% \midrule[1.0pt]
\textbf{Original (Neg)} & \textbf{Vanilla Transferred (Pos)} & \textbf{VAE Transferred (Pos)}\\
\midrule[0.1pt]
crap service with mediocre food is not a good business model to live by . & good service and the food is delicious . & good service with delicious food , good business model to live by . \\
\midrule[0.1pt]
this is a horrible representation of a deli .	 & this is a great place to go in this area .  & this is a good place of a deli .\\
\midrule[0.1pt]
the staff does a horrible job  with my teenagers .	& the staff works well with my teenagers . & the staff does a great job working with my teenagers .\\
\bottomrule[1.0pt]
\end{tabular}
}
\end{center}
\caption{Examples of sentiment polarity control.}
\label{tab:morecasemg}
\end{table*}%

\begin{table*}[!t]
\begin{center}
\resizebox{\linewidth}{!}{
\begin{tabular}{p{6cm}p{6cm}p{6cm}}
\toprule[1.0pt]
\textbf{Original (Now)} & \textbf{Vanilla Transferred (Past)} & \textbf{VAE Transferred (Past)}\\
%i will not be purchasing more of these as there are other toys that they like better . &i was not expecting a toy of that at all about that to be as good as . &i tried my other ones as a gift to someone who loves to play UNK the UNK .\\
\midrule[0.1pt]
this machine is exactly what the name says it is - a speller . & this machine was exactly a speller . & The machine was a speller, just as its name indicated .\\
\midrule[0.1pt]
it's so small (of course) and it's really only good for nuts . &  it was so small and only good for nuts . &it was very small and only useful for nuts in the past , just as it is now .\\
%\midrule[0.1pt]
%this soup is my favorite progresso soup, so my rating is for amazon's handling .  & this soup was my favorite progresso soup &  this soup was my favorite progresso soup in the past, and my rating is based on how Amazon handled it.\\
% \midrule[1.0pt]
\midrule[1.0pt]
\textbf{Original (Past)} & \textbf{Vanilla Transferred (Future)} & \textbf{VAE Transferred (Future)}\\
\midrule[0.1pt]
i did not like the taste of this at all. &  i will never like this taste . &   i will never like this taste any more .\\
\midrule[0.1pt]
i was not impressed, but at least i tried.	&  I will never be impressed . &I will not be impressed, but at least I will try.\\
\midrule[1.0pt]
% \midrule[1.0pt]
\textbf{Original (Future)} & \textbf{Vanilla Transferred (Past)} & \textbf{VAE Transferred (Past)}\\
\midrule[0.1pt]
i'm going to e-mail the company but in the meantime, if you drink this tea, stop. &  I emailed the company  . &I emailed the company, stop drinking this tea . \\
\midrule[0.1pt]
i'm probably going to end up throwing all of these out . & I threw all this out probably . & I probably ended up throwing all of these out.\\
\bottomrule[1.0pt]
\end{tabular}
}
\end{center}
\caption{Examples of tense control.}
\label{tab:morecasetense}
\end{table*}%

\begin{table*}[!t]
\begin{center}
\resizebox{\linewidth}{!}{
\begin{tabular}{p{6cm}p{6cm}p{6cm}}
\toprule[1.0pt]
\textbf{Original } & \textbf{Vanilla Transferred } & \textbf{VAE Transferred}\\
%i will not be purchasing more of these as there are other toys that they like better . &i was not expecting a toy of that at all about that to be as good as . &i tried my other ones as a gift to someone who loves to play UNK the UNK .\\
\midrule[0.1pt]
My friend and I have been having good fun queuing Mei/Winston, trololo, and 2 bottles of whisky. (Joy) &  My friend and I have been anxiously queueing Mei/Winston, trololo, and 2 bottles of whisky  (fear) & My friend and I anxiously wait in the endless, crawling queue for Mei/Winston, trololo,  and 2 bottles of whisky. (fear)\\
\midrule[0.1pt]
And then you're stuck on that medication for life, sounds like a scary trade-off. (fear) &And then I guess you're stuck taking that medicine forever and ever, seems like a real bummer of a trade-off.(sadness) &And then you're frustratedly stuck on that medication for life, sounds like a grim trade-off. (sadness)\\
\midrule[0.1pt]
it's hurting my heart :( why did you leave me (sadness) &It's sickening my heart :( why did you abandon me. (disgust) &It's breaking my heart :( why did you betray and leave me in such a repulsive manner. (disgust)\\ 
\midrule[0.1pt]
Gross. I hate using those reusable cloth ones because they retain smells. You've got cats? I know, I can smell them.  (disgust) &Damn it. It infuriates me to use those reusable fabric ones since they hold onto scents. Throw out these cats! (anger) &Damn it. I detest using those reusable cloth ones because they retain odors. You've got cats? I know, I can detect their stench. (anger)\\
\midrule[0.1pt]
I do remember this and wanted nothing more than to kill that guy (anger) &Yeah I totally remember this and was like, woah, my mind was blown about that guy. (surprised) &I do recall this vividly and was utterly astonished at that guy. (surprised)\\
 
\midrule[0.1pt]
Can't believe [NAME] has been in the league that long... (surprised) & It's amazing that [NAME] has been in the league that long! (joy) &It's so thrilling that [NAME] has already been playing in the league for that many years! (joy)\\ 
\bottomrule[1.0pt]
\end{tabular}
}
\end{center}
\caption{Examples of Ekman control in GoEmotions dataset.}
\label{tab:morecasetense_goemo}
\end{table*}%

\subsection{Human Evaluation}
We also conducted a human evaluation for the attribute control results. We sampled 1,000 examples from each of Yelp and Amazon, and changed their attribute value to the opposite value (``Positive''$\rightarrow$``Negative'', ``Nega\-ti\-ve''$\rightarrow$``Positive''). Then, we collected the generated sentences and asked 3 data graders to give a score to the sentences on 3 metrics (transfer accuracy (TA), content preservation (CP), and language quality (LQ)). Among them, TA is a percentage, CP and LQ are scored between $1\sim 5$. The detailed questions are listed in the appendix. We randomly shuffled  the sentences to remove the ordering hint. The final result of human evaluation is shown in Table~\ref{tab:he}. The inter-rater agreements
(the Krippendorff’s alpha values~(\citeyear{krippendorff2004content})) of the three metrics are $0.84$, $0.89$, and $0.92$, all of them are acceptable due to  Krippendorff’s
principle~(\citeyear{krippendorff2004content}). We can see that  our proposed method \Mname\ outperforms the baseline in each of the human evaluation metrics. We also listed some generated cases in Appendix~\ref{sec:casestudy}.
   
\begin{table}[!ht]
\centerline{
\resizebox{\linewidth}{!}{
\begin{tabular}{llccc}
\toprule[1.0pt]
 && TA & CP & LQ\\
 \midrule[0.5pt]
\multirow{4}{*}{Yelp}
&  \cite{logeswaran2018content} & 86.01 & 3.81 & 3.89 \\
&\cite{lample2019multipleattribute} & 82.32  &3.59   & 4.28 \\
&\cite{john-etal-2019-disentangled}(VAE) & 85.89 & 3.65  & 4.25\\
&MTDNA (Vanilla) &84.28 & 3.69& 4.32\\ 
&MTDNA (VAE) & 86.04& 3.78&4.39\\ 
&\cite{qian-etal-2022-controllable} & 83.43& 3.65&4.41\\ 
&\Mname (Vanilla) & 85.42 & 3.70 &4.32 \\ 
&\Mname (VAE) & \textbf{87.98} &\textbf{3.90} &\textbf{4.43} \\ 
 \midrule[0.5pt]
\multirow{4}{*}{Amazon}
&  \cite{logeswaran2018content} & 80.21 &  3.68  &  3.73  \\
&\cite{lample2019multipleattribute} & 77.76    &3.14   & 3.66 \\
&\cite{john-etal-2019-disentangled}(VAE) &  82.23   & 3.27  & 3.75\\
&MTDNA (Vanilla) &79.03 & 3.34& 3.74 \\ 
&MTDNA (VAE) & 83.28& 3.52&4.08\\ 
&\cite{qian-etal-2022-controllable}& 80.75& 3.21&4.10\\ 
&\Mname (Vanilla) &80.56  & 3.68 & 3.76\\ 
&\Mname (VAE) &\textbf{83.96}  &\textbf{3.75} &\textbf{4.32} \\ 
\bottomrule[1.0pt]
\end{tabular}
}}
\caption{Human evaluation results on  Yelp and Amazon. }
%\vspace*{1ex}
\label{tab:he}  
\end{table}%

\section{Discussion}

Recent work has explored utilizing large language models (LLMs) like ChatGPT and GPT-4 for controllable text generation. For example, \newcite{reif2021recipe} have proposed methods to steer text style transfer in these LLMs by conditioning on discrete attributes or continuous latent representations. Compared to our approach, a key difference is that we train our model end-to-end to disentangle latent attributes, while LLMs rely on prompting or fine-tuning approaches applied post-hoc.

While promising, utilizing LLMs for attribute-controlled generation remains challenging. The discrete prompting approach can yield brittle or superficial style changes, as the models' understanding of prompted attributes is imperfect and limited to correlation patterns in the pretraining data~\cite{reif2021recipe,luo2023prompt}. Latent space steering has shown more coherent style transfer, but current methods rely on complex optimization schemes or assume access to an attribute classifier~\cite{john-etal-2019-disentangled,sha2021multi}. In contrast, our model learns disentangled representations directly from data through closed-loop contrastive training.

\section{Limitations}
\paragraph{Controlling the attribute's intensity}

Our model is not designed to control the  intensity of an attribute, like generating  some neutral sentence instead of ``pos'' or ``neg''. If we want to generate a neutral sentence anyway, we just need to take the average vector of the mean value of the ``pos'' and ``neg'', and replace the original semantic style vector. Then, the decoder will generate a neutral sentence. However, this method will not always be successful,  because there is no guarantee that these latent spaces are smoothly distributed with overlapping regions, and the decoder may not have been required to generate such texts with novel style features during training. To better control the attribute's intensity, it is required to design some special mechanics in a supervised manner.

\paragraph{Difficult attributes}

Apart from the simple text attributes, there are also some complex attributes like some specific author's style of writing, which are usually intertwined together in the latent space. Discrete categorical style types are hard to design for such kind of complex attributes. Whether disentanglement can be used for controlling complex attributes requires further research.

\section{Conclusion}
In this paper, we proposed a novel explicit disentanglement method, called 
contrastive learning disentanglement (\Mname), which uses contrastive learning as the core method. Differently from previous works, we re-disentangle the reconstructed sentences, and conduct contrastive learning between the disentangled vectors and the re-disentangled vectors. To encourage the disentanglement of the  attributes' latent space, we propose the re-disentangled contrastive loss $L_\text{re}$ and the transferred  re-disentangled contrastive loss $\tilde L_k$. The latter fully imitates the attribute control process. To encourage  content preservation, we proposed the content contrastive loss $L_c$, which contains three sub-losses. These sub-losses make the content space more distinguishable and encourage the content keep unchanged during attribute control. Our proposed method is not only much easier in the mathematical derivations, it also   outperforms all the compared methods in the evaluation metrics according to our experimental results.

\section*{Acknowledgement}
This work was supported by the ESRC grant ES/S010424/1 “Unlocking the Potential of AI for English Law”, by the National Natural Science Foundation of China under grant No.\ KZ37117501, No.\ ZG216S23E8, and  No.\ KZ60030101, by the Alan Turing Institute under the EPSRC grant EP/N510129/1, 
and by the AXA Research Fund. We also acknowledge the use of Oxford’s Advanced Research Computing
(ARC) facility, of the EPSRC-funded Tier 2 facility JADE (EP/P020275/1), and of GPU computing support by Scan Computers
International Ltd.

\bibliography{cite}
\bibliographystyle{acl_natbib}

\onecolumn

\appendix
\section*{Appendix}
\section{Human Evaluation Questions}
\subsection{Transfer Accuracy (TA)}
  
Q: Do you think the given sentence belongs to positive sentiment or negative sentiment?

\begin{itemize}
\item A: Positive. 
\item B: Negative.
\end{itemize}

\subsection{Content Preservation (CP)}

Q: Do you think the generated sentence has the same content with the original sentence, although the sentiment/tense is different?

Please choose a score according to the following description. Note that the score is not necessary to be integer, you can give scores like $3.2$ or $4.9$ by your feeling.
\begin{itemize}
\item 5: Exactly. The contents are exactly the same.
\item 4: Highly identical. Most of the content are identical.
\item 3: Half. Half of the content is identical.
\item 2: Almost Not the same.
\item 1: Totally different.
\end{itemize}

\subsection{ Language Quality (LQ)}

Q: How fluent do you think the generated text is? Give a score based on your feeling.

Please choose a score according to the following description. Note that the score is not necessary to be integer, you can give scores like $3.2$ or $4.9$ by your feeling.
\begin{itemize}
\item 5: Very fluent. 
\item 4: Highly fluent. 
\item 3: Partial fluent. 
\item 2: Very unfluent. 
\item 1: Nonsense.
\end{itemize}

\section{Detailed Model Structure}
The encoder and decoder in this paper are in the LSTM architecture. Given an input sentence $X=\{x_1,\ldots, x_n\}$, the representation of the sentence $z$ is computed by:
\begin{equation}
    z=LSTM(x_1,\ldots, x_n)\,,
\end{equation}
where $z$ is the output of the last LSTM cell. 

In the vanilla autoencoder, $z$ is split into a style vector and a content vector by a feed-forward layer as follows:
\begin{equation}\label{eq:split}
    [s,c] = \tanh(W_hz+b_h)\,,
\end{equation}
where $W_h$ and $b_h$ are trainable parameters. 

In the variational autoencoder, the output of the encoder is further mapped into two vectors $\mu$ and $\log(\sigma^2)$. The latent vector $z$ is sampled from the Gaussian distribution $\mathcal N(\mu,\sigma)$. The style vector and content vector are split by Eq.~\eqref{eq:split}.

After the transfer of the style vector ($s$ changed to $s'$), the style vector and content vector  are merged into a new latent vector $z'$, which is the input to the LSTM decoder  as the initial state.

\section{Detailed Model Settings}
The encoder and decoder are set as 2-layer LSTM RNNs with input dimension of 100, and the hidden size is 150. The hyperparameters are set to $\lambda_\text{ori}=1.0, \lambda_\text{re}=1.0, \lambda_k=1.5, \lambda_c=2.0, \lambda_\text{KL}=0.01$ and $\tau=100$. We used Adam for optimization, and the learning rate is set to $0.001$. The model is run on an Nvidia v100 GPU. 
\end{document}